\documentclass{article}

\usepackage[preprint]{neurips_2024}

\usepackage[utf8]{inputenc} 
\usepackage[T1]{fontenc}    
\usepackage{hyperref}       
\usepackage{url}            
\usepackage{booktabs}       
\usepackage{amsfonts}       
\usepackage{nicefrac}       
\usepackage{microtype}      
\usepackage{xcolor}         

\usepackage[ruled,vlined]{algorithm2e}
\usepackage{amssymb}

\usepackage{microtype}      
\usepackage{xcolor}         
\usepackage{graphicx}
\usepackage{amsmath}
\usepackage{wrapfig}
\usepackage{enumitem}
\usepackage{amsthm}
\usepackage{multirow}
\usepackage{array} 
\usepackage{multicol}
\usepackage{lipsum}
\usepackage{subcaption}
\usepackage{algorithmic}

\newtheorem{theorem}{Theorem}

\theoremstyle{definition}

\newtheorem{assumption}{Assumption}

\theoremstyle{remark}

\title{Towards Client Driven Federated Learning}

\author{%
  Songze Li \\
  Southeast University\\
  \texttt{songzeli@seu.edu.cn} \\
  \And
  Chenqing Zhu \\
  Hong Kong University of Science and Technology (Guangzhou) \\
  \texttt{czhu032@connect.hkust-gz.edu.cn} \\
}

\begin{document}

\maketitle

\begin{abstract}
Conventional federated learning (FL) frameworks follow a server-driven model where the server determines session initiation and client participation, which faces challenges in accommodating clients' asynchronous needs for model updates. We introduce Client-Driven Federated Learning ({\ttfamily CDFL}), a novel FL framework that puts clients at the driving role. In {\ttfamily CDFL}, each client independently and asynchronously updates its model by uploading the locally trained model to the server and receiving a customized model tailored to its local task. The server maintains a repository of cluster models, iteratively refining them using received client models. Our framework accommodates complex dynamics in clients' data distributions, characterized by time-varying mixtures of cluster distributions, enabling rapid adaptation to new tasks with superior performance. In contrast to traditional clustered FL protocols that send multiple cluster models to a client to perform distribution estimation, we propose a paradigm that offloads the estimation task to the server and only sends a \emph{single} model to a client, and novel strategies to improve estimation accuracy. We provide a theoretical analysis of {\ttfamily CDFL}'s convergence. Extensive experiments across various datasets and system settings highlight {\ttfamily CDFL}'s substantial advantages in model performance and computation efficiency over baselines.
\end{abstract}

\section{Introduction}

\label{introduction}

Federated Learning (FL) \citep{mcmahan2017communication} is a distributed learning framework that allows for collaborative training of a global model across multiple clients while keeping their raw data local. 
In nearly all current FL frameworks, the central locus of control invariably resides with the server. That is, the server initiates training sessions for model update, and determines which clients should participate and when. However, this server-driven paradigm may fall short in serving clients' needs, especially in scenarios where clients experience asynchronous changes of their data distributions. Specifically, a client experiencing a concept drift will suffer from sub-optimal performance using the old model, until the server calls for the next model update.


In this paper, we attempt to address the above challenge by proposing a novel \textbf{C}lient-\textbf{D}riven Federated Learning ({\ttfamily CDFL}) framework, which \textit{empowers each individual client to assume a more autonomous role in the FL process}. 
Specifically, we focus on the scenario where each client collects data from a mixture of distributions. As the mixing ratio varies over time, the client may seek help from the server, who acts as a service provider, in updating its local model to match the new distribution. As a real-life example, consider a skincare maintenance application, where users' skin types exhibit complexity — perhaps featuring a combination of oiliness and dryness in different areas of skin, reflecting a mixture of distributions. Additionally, it is common for users' skin conditions to vary with seasons, leading to shifts in distributions. Another example is a retail chain with various branches, each of which sell commodities of different store categories. The commodities offered by these branches may evolve based on changing of customer preferences, creating a dynamic mixture of distributions. In {\ttfamily CDFL}, in sharp contrast to conventional server-driven paradigm, each client possesses complete autonomy in deciding when to update its model, and the servers plays a passively assistive role for the clients to adapt to their new distributions.

\begin{wrapfigure}{r}{0.5\textwidth}
    \centering
    \includegraphics[width=0.5\textwidth]{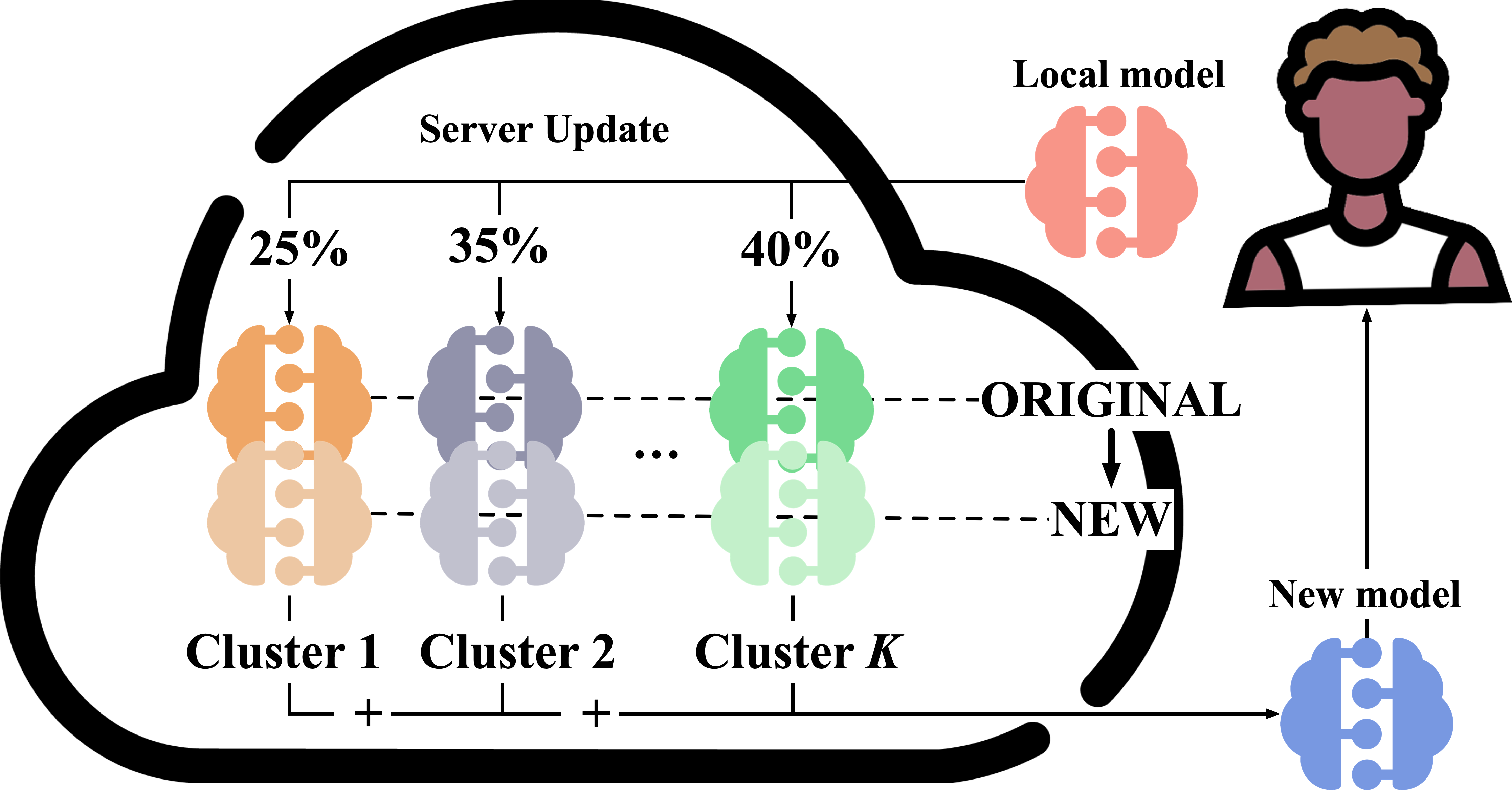}
    \caption{High-level view of {\ttfamily CDFL}.}
    \label{pic: high level view}
\end{wrapfigure}

To tackle clients' varying data distributions, we adopt the clustered FL setting where $K$ base cluster models are maintained at the server~\citep{sattler2020clustered, sattler2020byzantine} to update clients' models. In existing clustered FL works, a crucial consideration is to measure the data distributions of clients. Many works distribute all cluster models to clients, leaving it to clients to determine the distribution based on local empirical loss \citep{ghosh2020efficient, mansour2020three, ruan2022fedsoft}. However, such an approach poses several challenges. Firstly, it places a significant communication burden to send all the cluster models; Secondly, it imposes substantial computational demands on clients, requiring them to calculate losses for each cluster and make comparisons. Some other approaches leverage distances between uploaded models to form client groups \citep{duan2021fedgroup}, imposing impractical synchronization requirements on clients for data uploads. In sharp contrast, as illustrated in Figure \ref{pic: high level view}, {\ttfamily CDFL} \emph{assigns the task of evaluating client data distribution to the server}. Based on the model uploaded by a client, the server analyzes its data distribution, and updates the cluster models. Subsequently, the server generates a personalized model and sends it to the client. This significantly simplifying clients' communication and computation compared with previous clustered FL solutions, reflecting another critical aspect of client-driven FL: ensuring good performance on clients with lightweight operations.

In the context of above clustered FL, and building upon the client-driven spirit, we develop an asynchronous {\ttfamily CDFL} framework that focuses on maximizing clients' performance and minimizing clients' complexity. Specifically, we introduce an effective newcomer cold start mechanism, a feature conspicuously absent in the majority of related works \citep{duan2021fedgroup, zeng2023stochastic}. Furthermore, our framework exhibits adaptability in addressing client distribution drift, a challenge specifically addressed in only one previous study \citep{duan2021flexible} within the context of clustered FL. \emph{{\ttfamily CDFL} is the first clustered FL framework that focuses on clients' autonomy, performance, and efficiency}. Compared with existing clustered FL works, the computation overhead of a {\ttfamily CDFL} client remains minimal: it only conducts the minimal required local model training for any FL protocol; the client's communication overhead is also minimized, with solely uploading and downloading \emph{one single model}, at any time completely determined by the client.

We provide convergence analysis that theoretically validates the convergence of client models on cluster and local tasks. Extensive experiments over different datasets and network settings attest to the substantial improvements of {\ttfamily CDFL} on cluster and client accuracies. Additionally, it significantly alleviates both communication and computational costs at clients over baselines.

\section{Related Work}

\textbf{Clustered Federated Learning (clustered FL)}. Hard clustering algorithms assume clients in the same group have identical data distribution \citep{briggs2020federated, ghosh2020efficient, mansour2020three}; while soft clustering methods assume the data of each client follows a mixture of multiple distributions \citep{ruan2022fedsoft, li2021federated}. In most cases, expectation-maximization (EM) methods are used to compute clients' distribution \citep{long2023multi, ma2022convergence, ghosh2022improved}, and global updates leverage methods based on {\ttfamily FedAvg} \citep{briggs2020federated}. Some works add proximal terms on clients' objectives for personalization \citep{tang2021personalized}. 

\textbf{Asynchronous Federated Learning (asynchronous FL)}. Asynchronous FL operates on resource-constrained devices \citep{xu2021asynchronous}. In typical asynchronous setups, the central server conducts global aggregation immediately upon receiving a local model \citep{xie2019asynchronous, wang2022asyncfeded, chen2020asynchronous}, or a set of local models \citep{nguyen2022federated, wu2020safa}. These asynchronous clients may be grouped into tiers for updating based on factors like staleness or model similarities \citep{park2021sageflow, wang2022asynchronous}, referred to as semi-asynchronous. However, this clustering typically contributes to a single global model, and sometimes, the server still selects the clients \citep{zhang2021csafl}. Existing clustered FL frameworks primarily operate within a synchronous setting. In the context of asynchronous FL, clients are sometimes grouped only to control staleness. 
Our framework is the first, to the best of our knowledge, to integrate clustered FL within an asynchronous setting.

\textbf{Personalized Online FL}. Mainstream online FL approaches primarily address the communication and computation burden on clients, with the goal of ensuring efficient FL training sessions, irrespective of whether the system operates synchronously or asynchronously \citep{jiang2023computation, zheng2023federated, gauthier2023asynchronous, damaskinos2022fleet}. Other research efforts in this domain focus on assisting clients in adapting to new task streams \citep{gogineni2022communication, ganguly2023multi, m2022personalized}. \cite{chen2020asynchronous} is also an asynchronous FL framework which accomplishes personalization by aggregation of local and global models. \cite{yu2021fedhar} proposes a personalized FL system based on human activity recognition (HAR), albeit with limited applicability to other data types. Additionally, \cite{deng2020adaptive} introduces an adaptive personalized method that combines local and global models, but overlooks potential asynchrony issues. In light of these observations, our proposed framework addresses personalized online FL challenges within an asynchronous setting by leveraging clustered FL algorithms.

\textbf{User-centric FL Frameworks}.
Few works have studied FL from a comprehensive user's perspective. \cite{mestoukirdi2021user, mestoukirdi2023user} claim to be user-centric, but are indeed personalized FL frameworks dealing with communication burdens. In \cite{khan2023pi}, the authors point out that existing FL works take away clients’ autonomy to make decisions themselves, and propose a token-based incentive mechanism that rewards personalized training. However, this work fails to consider the asynchrony among clients, making it insufficient to provide full autonomy to clients. Note that the shift in clients' distribution is distinct from Federated Continual Learning (FCL)~\cite{yoon2021federated}, which primarily aims to minimize catastrophic forgetting. Our focus lies solely in enabling clients to seamlessly adapt their models to new data during distribution shifts.

\section{Problem Definition}
Consider an FL system with one central server and many distributed clients. The server maintains $K$ cluster models, corresponding to $K$ distributions $P_1,\ldots,P_K$, and has a proxy dataset $D_k$ for each $P_k$. Note that the existence of small-size proxy datasets is rather a mild and common assumption in FL literature \citep{wang2024transtroj, li2019fedmd, lin2020ensemble}. The value of $K$ is determined a priori, according to the type of service (e.g., genders or ethnicities in the skincare service), or is deducted from a small amount of data collected in advance. Given a loss function $l(w; x, y)$, each cluster $k \in [K] \triangleq \{1,\ldots,K\}$ aims to find an optimal model $w_k$ that minimizes the objective 
\begin{equation}
     F_k(w_k) = \mathbb{E}_{(x, y) \sim P_k}[l(w_k; x, y)].
    \label{cluster objective}
\end{equation}

The training takes $T$ global epochs. For each epoch $t \in [T]$, some client $m$ collects local data following a mixture of distribution $P_m^t = \sum_{k=1}^K \alpha^t_{mk}P_k$, with $\alpha^t_{mk} \in [0, 1] $ and $\sum_{k=1}^K\alpha_{mk}^t = 1$. Here $\alpha^t_{mk}$ is the importance weight of cluster $k$ to client $m$ at epoch $t$. The importance weights may vary over time, i.e., $\alpha_{mk}^t \neq \alpha_{mk}^{t'}$ for $t \neq t'$, and are unknown to the client. Each time when client $m$'s data distribution shifts, it may choose to fit the local model $v_m^t$ to the new distribution, and uploads it to the server. The server enhances $v_m^t$ to construct a personalized model $u_m^t$ for client $m$:
\begin{align}
    u_m^t = g(v_m^t, \{w_k^{t-1}\}_{k=1}^K, \{D_k\}_{k=1}^K),
\end{align}
following some rule $g$, and returns $u_m^t$ to client $m$. In the meantime, server also updates the cluster models using some function ${\bf h}$, such that
\begin{align}
    (w_1^t,\ldots,w_K^t)= {\bf h}(v_m^t, \{w_k^{t-1}\}_{k=1}^K, \{D_k\}_{k=1}^K).
\end{align}

Specifically, the local model $v_m^t$ is obtained by optimizing the local objective
\begin{equation}
    J_m^t(v_m^t; u_m^\tau) = 
     \frac{1}{n_m^t} \mathbb{E}_{(x^i, y^i)\sim P_m^t} \big[\sum_{i=1}^{n_{m}^t} l(v_m^t; x^i, y^i)\big] + \frac{\rho}{2} \left\|v_m^t-u_m^\tau\right\|^2.
    \label{client objective}
\end{equation}
Here $n_m^t$ is the number of data samples at client $m$ in epoch $t$; $\rho$ is some scaling parameter; $\tau < t$ is the last epoch when client $m$ uploads its model $v_m^\tau$ to the server. 

\begin{figure*}[ht!]
  \begin{center}
    \includegraphics[width=\textwidth]{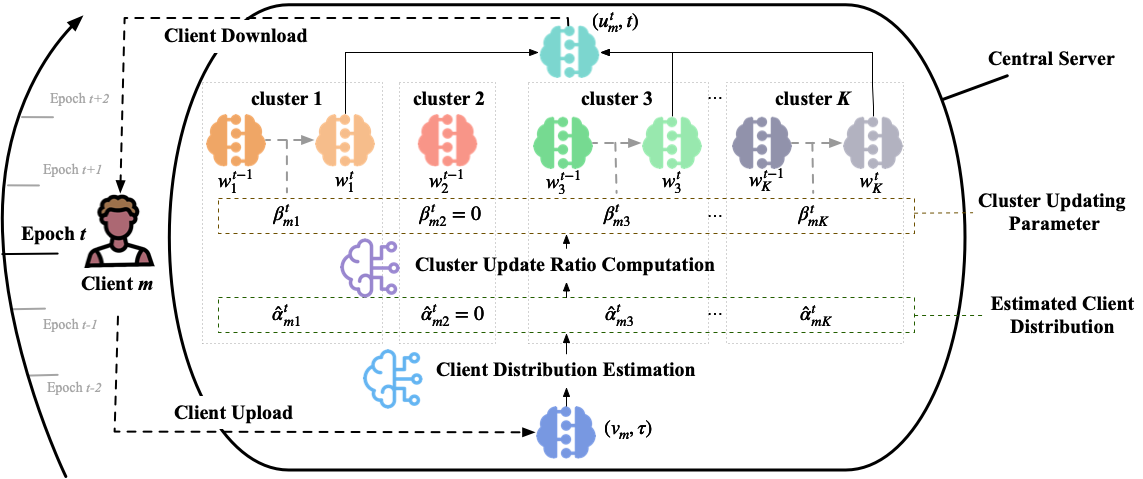}
    \caption{{\ttfamily CDFL} workflow. Client $m$ uploads model and epoch index of last model update $(v_m, \tau)$ to the server. Server performs distribution estimation, derives cluster updating parameters to update the cluster models, and finally constructs an aggregated model $u_m^t$ to send back to the client. Note here as the client's distribution is estimated not to contain $P_2$, cluster 2's model is not updated and not used in computing $u_m^t$.}
    \label{pic:CDFL workflow}
  \end{center}
\end{figure*}

\section{Client-driven Federated Learning}
\label{client-driven Federated Learning}
We describe the operations of the client and the server respectively in the proposed {\ttfamily CDFL} framework. The entire workflow of {\ttfamily CDFL} is depicted in Figure \ref{pic:CDFL workflow} and detailed in Algorithm \ref{alg:CDFL}.

\begin{algorithm}[ht!]
	\caption{{\ttfamily CDFL}}
        \label{alg:CDFL}
	\KwIn{Server pre-trained model $w_k^0$, server proxy dataset $D_k\sim P_k$ ($k \in [K]$), staleness threshold $\tau_0 < T$, cluster update threshold $\beta_0 \in (0, 1)$}
	\KwOut{Local model parameter $v_m$, cluster model parameters $w_1,\ldots,w_K$} 

    \textbf{Initialization}: Server sends $\left(u^0, 0\right)$ to each client, $u^0 = \frac{1}{K}\sum_{k=1}^K w_k^0$. Global epoch $t \leftarrow 0$. Run {\ttfamily Client()} thread and {\ttfamily Server()} thread asynchronously in parallel.
    
\SetKwFunction{FMain}{Server}
\SetKwProg{Fn}{Thread}{:}{}
\Fn{\FMain{}}{

    \While{no client uploads}
    {
        Wait for client update. \If{client $m$ uploads $\left(v_m, \tau\right)$}{
            $t \leftarrow t+1$; $u_m^t \leftarrow$ {\ttfamily ServerUpdate} ($v_m, \tau, t$); send $\left(u_m^t, t\right)$ to client $m$.
        }}}	
    \SetKwFunction{FMain}{Client}
    \SetKwProg{Fn}{Thread}{:}{}
    \Fn{\FMain{}}{
        \ForEach{client $m$ in parallel}{
            Receive pair $\left(u_m, t\right)$ from server. Set local model $v_m\leftarrow u_m $, local timestamp $t_m \leftarrow t$.
            
            \While{active}{ 
                \If{choose to upload}{
                    Define $J_m(v_m; u_m)$ as in \eqref{client objective}.
                    
                    \ForEach{local iteration $h$ } {
                        $v_{m, h}\leftarrow v_{m, h-1} -\gamma\nabla J_m(v_{m, h-1}; u_m)$
                        
                        \tcc{learning rate $\gamma$}
                    }
                    Upload ($v_m, t_m$) and wait for server response.
                }
            }
        }
    }
    \SetKwFunction{FMain}{ServerUpdate}
    \SetKwProg{Fn}{Function}{:}{}
    \Fn{\FMain{$v_m, \tau, t$}}{
        \If{$t-\tau > \tau_0$}{
            \tcc{Client deprecated. Do not update cluster models.}
            
            \textbf{foreach} $k \in [K]$ \textbf{do} $w_k^t \leftarrow w_k^{t-1}$
            
            \textbf{return} $u_m^t=\sum_{k=1}^K \hat{\alpha}_{mk}^\tau w_k^{t}$.
        }
        $\hat{\alpha}_{m1}^t, ..., \hat{\alpha}_{mK}^t \leftarrow $  {\ttfamily DistributionEstimate}
        ($v_m, w_1^{t-1}, ..., w_K^{t-1}, D_1, ..., D_K, t$)
        
        $\beta_{m1}^t, ..., \beta_{mK}^t \leftarrow $ {\ttfamily UpdateRatioCompute} 
        ($\hat{\alpha}_{m1}^t, ..., \hat{\alpha}_{mK}^t, \beta_0, \tau, t$)
        
        \textbf{foreach} $k \in [K]$ \textbf{do}
            $w_k^{t}\leftarrow\left(1-\beta_{mk}^t\right)w_k^{t-1}+\beta_{mk}^t v_m$
            
        \textbf{return} $u_m^t=\sum_{k=1}^K \hat{\alpha}_{mk}^t w_k^{t}$.}
\end{algorithm}

\subsection{Client Update} 

\textbf{Initialization.} {\ttfamily CDFL} is designed to be fully open and dynamic, which does not presuppose a fixed total number of clients. A client can seamlessly joint the system, simply via retrieving an initialization tuple from the server, comprising an initial model and the index of current epoch, denoted as $(u_m^t, t)$.

\textbf{Training and Uploading.} Whenever a client $m$ feels necessary to perform a model update, perhaps due to the occurrence of distribution shift and the resulting performance degradation, it performs local training on current dataset to obtain a local model $v_m$, and uploads $(v_m, \tau)$ to the server, where $\tau$ is the index of the latest epoch in which client $m$ updates its model with the server. Having received $v_m$, the server generates an enhanced model $u_m^t$ for current epoch $t$, and returns $(u_m^t,t)$ to client $m$. Finally, client $m$ updates $\tau = t$, and starts to use $u_m^t$ for local tasks. 

\subsection{Server Update}

Throughout the entire process of {\ttfamily CDFL}, the server passively waits for clients' update requests. Upon receipt of an uploaded model, the server first increments the epoch counter to obtain the current epoch $t$, then the server initiates a two-step evaluation process. Firstly, it checks if the client's model is too stale. Specifically, as client $m$ uploads $(v_m, \tau)$, if $t-\tau>\tau_0$ for some preset staleness threshold $\tau_0$, the server refrains from updating the cluster models, i.e., $w_k^t = w_k^{t-1}$ for all $k$, and returns the client a personalized model as an aggregation of current cluster models. Otherwise, the server proceeds to estimate client $m$'s data distribution. Subsequently, it updates each cluster model, and constructs a new personalized model for the client, as an aggregation of the updated cluster models. 

\textbf{Distribution Estimation.} Upon client $m$ uploading $v_m$ at epoch $t$ (referred to as $v_m^t$ for clarity), the estimation of its distribution hinges on several components, including $v_m^t$, the latest cluster models $w_1^{t-1},\ldots,w_K^{t-1}$, and their proxy datasets $D_1,\ldots,D_K$. Two key metrics are evaluated to estimate the importance weights in client $m$'s distribution. The first is the empirical loss of $v_m^t$ on $D_k$, denoted by $F(v_m^t; D_k)$, for all $k \in [K]$. Specifically, when $F(v_m^t; D_k) < F(v_m^t; D_{k'})$, $v_m^t$ fits $P_k$ better than $P_{k'}$, which indicates that cluster $k$ should have a higher importance weight than cluster $k'$ in client $m$'s distribution $P_m^t$. The second metric is a measure on the similarity between $v_m^t$ and the cluster models. This similarity can be materialized either through the loss difference between the cluster model and the client's model on proxy data, i.e., $|F(w_k^{t-1}; D_k)-F(v_m^t; D_k)|$, or the $l_2$ distance between the models, i.e., $\left\|v_m^t-w_k^{t-1}\right\|$. 
Leveraging these metrics, we propose {\ttfamily DistributionEstimation} in Algorithm \ref{alg:DistributionEstimation & UpdateRaTioCompute} to estimate the importance weights of $P_m^t$ as $\hat{\alpha}_{m1}^t, \ldots,\hat{\alpha}_{mK}^t$.

\textbf{Cluster Updating.} The server updates the model of each cluster $k$ as follows,
\begin{equation}
    w_k^t = (1-\beta_{mk}^t)w_k^{t-1}+\beta_{mk}^t v_m^t,
\end{equation}
where $\beta_{mk}^t$ is the updating ratio contributed by client $m$ to cluster $k$ at epoch $t$. The value of $\beta_{mk}^t$ depends on 1) the correlation between $v_m^t$ and $w_k^{t-1}$ as measured by the weight $\hat{\alpha}_{mk}^t$ evaluated for distribution estimation; and 2) the staleness of $v_m^t$ indicated by the epoch index $\tau$ in which client $m$'s model is lastly updated. Detailed procedures for computing the updating ratio are elucidated in {\ttfamily UpdateRatioCompute} in Algorithm \ref{alg:DistributionEstimation & UpdateRaTioCompute}.

\vspace{-3mm}
\subsection{Convergence Analysis}
\vspace{-2mm}
\label{convergence analysis}
To assist the convergence analysis of {\ttfamily CDFL}, we make the following assumptions that are standard in analyses of asynchronous and clustered FL algorithms (see, e.g.,~\cite{xie2019asynchronous,ghosh2020efficient,ruan2022fedsoft}).
\begin{assumption}
$F_k$ is $L_k$-smooth and $\mu_k$-strongly convex and for some $L_k, \mu_k > 0$, for all $k \in [K]$.
\label{assumption1}
\end{assumption}
\begin{assumption}
Each client executes at least $H_{min}$ and at most $H_{max}$ local model updates before uploadin to server. 
\label{assumption2}
\end{assumption}
\begin{assumption}
In \eqref{client objective}, we simplify $v_m^t$ as $v$ denoting local model of client $m$ at current epoch $t$, and $u_m^\tau$ as $u$ denoting the latest model received from server. Let $f_m^t(v) \triangleq \frac{1}{n_m^t} \mathbb{E}_{(x^i, y^i)\sim P_m^t} [\sum_{i=1}^{n_m^t} l(v; x^i, y^i)]$, then $J_m^t(v; u) = f_m^t(v)+\frac{\rho}{2}\left\|v-u\right\|^2$. We assume $\forall m$, and $\forall t \in T$, we have $\left\|\nabla f_m^t(v)\right\|^2 \leq V_1$ and $\left\|\nabla J_m^t(v; u)\right\|^2 \leq V_2$, for some constants $V_1$ and $V_2$.
\label{assumption3}
\end{assumption}
\begin{assumption}
The distance between optimal models of different clusters is bounded as $a_0\Delta \leq \left\|w_k^*-w_{k'}^*\right\|\leq \Delta$, for some constant $\Delta$, and $0 \leq a_0 \leq 1$, for all $k \neq k'$.
\label{assumption4}
\end{assumption}
\begin{assumption}
The $l_2$ norm of cluster $k$'s model $w_k$, $\forall k \in [K]$, is bounded as $\left\|w_k\right\| \leq a_k\Delta$, for some $a_k > 0$.
\label{assumption5}
\end{assumption}

\begin{theorem}
\label{theorem 1}
For a client with model $v$ and a cluster $k$, we consider consecutive $S_k$ epochs, such that in each epoch the data of the client contains an non-zero component of $P_k$, and the client and cluster $k$ updates $v$ and $w_k$ respectively as in Algorithm \ref{alg:CDFL}, then with the above assumptions, choosing $\rho  \geq \frac{2V_1+\frac{1}{2}\left\|v-u\right\|^2+\sqrt{4\left\|v-u\right\|^2(1+V_1)\epsilon}}{2\left\|v-u\right\|^2}$ for all possible $v$ and $u$, and a small constant $\epsilon > 0$, we have 
\begin{align*}
    & \qquad \mathbb{E}[\left\|\nabla F_k(v)\right\|^2] \leq \frac{\mathbb{E}[F_k(w_k^0)-F_k(w_k^{S_k})]}{\beta_0\gamma\epsilon S_k H_{min}} +\frac{\left(\frac{L_k}{2}+\rho H_{max}+\frac{\rho H_{max}^2}{2}\right)\gamma H_{max}V_2}{\epsilon H_{min}}\\
    &+\frac{\sqrt{V_1} \left(2\sum_{i=1}^K a_i+(2K+1)a_k+K\right)\Delta}{\gamma\epsilon H_{min}} +\frac{\left(\frac{L_k}{2}+\rho\right)\left(2\sum_{i=1}^K a_i+(2K+1)a_k+K\right)^2\Delta^2}{\gamma\epsilon H_{min}},
\end{align*}
where $w_k^0$ and $w_k^{S_k}$ denotes cluster $k$'s initial model and the model after $S_k$ updates respectively.
\end{theorem} 
\begin{proof}
See proof in Appendix~\ref{proof1}.
\end{proof}

This theorem demonstrates the convergence of a client's model on the loss function of a single cluster $k$. The upper bound on the gradient norm increases with $H_{max}$ and $\Delta$, and decreases with $H_{min}$. Intuitively, a larger $H_{min}$ indicates a more sufficient local training, and a smaller $H_{max}$ reduces the effect of overfitting; $\Delta$ represents the distance between different clusters, and larger distance leads to slower convergence. Since a client's local loss is a convex combination of the cluster losses, and the above theorem holds true for each cluster, it implies a good performance on the client's local task.

\begin{algorithm}[H]
    \caption{DistributionEstimation \& UpdateRatioCompute}
    \label{alg:DistributionEstimation & UpdateRaTioCompute}
    \SetKwFunction{FMain}{DistributionEstimation}
    \SetKwProg{Fn}{Function}{:}{}
    \Fn{\FMain{$v_m, w_1, \ldots, w_K, D_1, \ldots, D_K, t$}}{
        \ForEach{$k \in [K]$}{
            $l_k\leftarrow F(v_m; D_k)$; $d_{1k} \leftarrow |F(w_k; D_k)-l_k|$; $d_{2k} \leftarrow \left\|v_m-w_k\right\|$
            
            $l_k \leftarrow l_k-l_{bar}$; $d_{1k} \leftarrow d_{1k}-d_{1bar}$; $d_{2k} \leftarrow d_{2k}-d_{2bar}$
            
            \tcc{$l_{bar}$, $d_{1bar}$, $d_{2bar}$ are hyper-parameters to control the scale.}
            
            $\hat{\alpha}_{mk}^t \leftarrow \frac{1}{K-1}\cdot\left(c_1\cdot\frac{\sum_{i\neq k}l_i}{\sum_i l_i}+c_2\cdot\frac{\sum_{i\neq k}d_{1i}}{\sum_i d_{1i}}+(1-c_1-c_2)\cdot\frac{\sum_{i\neq k}d_{2i}}{\sum_i d_{2i}}\right)$
        
            \tcc{$c_1, c_2 \in [0, 1]$ are hyper-parameters. $c_1+c_2 \in [0, 1]$.}
        }
        $\hat{\alpha}_{m1}^t, \ldots, \hat{\alpha}_{mK}^t\leftarrow  \text{softmax}(\hat{\alpha}_{m1}^t\cdot A, ..., \hat{\alpha}_{mK}^t \cdot A)$
        
        \tcc{ $A > 0$ is the hyper-parameter to magnify the difference of estimation results of clusters. Estimated weights $\hat{\alpha}_{mk}^t \in [0, 1]$ for $k \in [K]$, and $\sum_{k=1}^K \hat{\alpha}_{mk}^t = 1$.}
    }
    \SetKwFunction{FMain}{UpdateRaTioCompute}
    \SetKwProg{Fn}{Function}{:}{}
    \Fn{\FMain{$\hat{\alpha}_{m1}^t, \ldots, \hat{\alpha}_{mK}^t, \beta_0, \tau, t$}}{
        \ForEach{$k \in [K]$}{
            $\beta_{11},\ldots, \beta_{1K} \leftarrow \hat{\alpha}_{m1}^t, \ldots, \hat{\alpha}_{mK}^t$; $\beta_{1max} \leftarrow \max ( \beta_{1k})$.
            
            \textbf{if} $\beta_{1k}< \beta_{1bar}$ \textbf{then} $\beta_{1k}\leftarrow0$; \textbf{else then} $\beta_{1k}\leftarrow \beta_{1k} / {\beta_{1max}}$; $e_k \leftarrow t$
            
            \tcc{$\beta_{1bar}$ is a preset threshold. Do not update cluster $k$ when $\beta_{1k} < \beta_{1bar}$. $e_k$ is the last epoch when cluster $k$ is updated.}
            
            \textbf{if} $e_k-\tau < b$ \textbf{then} $\beta_{2k} \leftarrow 1$; \textbf{else then} $\beta_{2k}\leftarrow 1/\left(a(e_k-\tau)+1\right)$
            
            \tcc{$a, b$ are hyper-parameters to control staleness.}
            
            $\beta_{mk}^t \leftarrow \beta_0\cdot\beta_{1k}\beta_{2k}$ \quad \tcc{$\beta_{mk}^t \in [0, \beta_0]$; $\beta_0$ governs the maximal local model modification to the cluster model.}
    }
    \textbf{return} $\beta_{m1}^t, ..., \beta_{mK}^t$
    }
\end{algorithm}

\section{Experiments}
\subsection{Setup}
\label{experiment setup}
We create FL clustered datasets via three public datasets: FashionMNIST \citep{xiao2017/online}, CIFAR-100 \citep{krizhevsky2009learning}, MiniImageNet-100 \citep{vinyals2016matching}. In order to simulate different
distributions, we augment the datasets using rotation, and create the Rotated FashionMNIST, Rotated CIFAR-100 and Rotated MiniImagenet-100 datasets. 
\textbf{Rotated FashionMNIST}: each cluster has 60,000 training images and 10,000 testing images with 10 classes; \textbf{Rotated CIFAR-100}: each cluster has 50,000 training images and 10,000 testing images with 100 classes; \textbf{Rotated MiniImagenet-100}: each cluster has 48,000 training images and 12,000 testing images with 100 classes. Each  dataset is applied by $i*\frac{360}{K} (i=0, ..., K-1)$ degrees of rotation to the images, resulting in $K$ clusters. We conduct experiments on various values of $K = 2,3,4, 6$.

We also perform a practical digit recognition task on three distinct datasets: the hand-written dataset composing of MNIST\citep{lecun2010mnist} and USPS\citep{hull1994database}; the Street View House Numbers (SVHN\citep{netzer2011reading}); and the lisence plate numbers collected from CCPD\citep{li2020ccpd}. The three datasets correspond to three clusters (or three different data distributions) in our setting. Consider an autonomous driving application where the client is a smart car that needs to recognize digits with its model on board, as the car traveling to different areas, the proportion of the digits from different types of images (digits from car plates or from house numbers) also changes, which is modeled by changing of mixing ratio of the three datasets in our experiment.
More details about the employed training datasets and models are provided in Appendix \ref{Model Structures.}. 

{\bf Baselines.} 1) {\ttfamily FedSoft-Async}. An asynchronous adaptation of the soft-clustering baseline \cite{ruan2022fedsoft}. Clients receive all cluster models from the server, and performs distribution estimation locally through identifying the model with the smallest loss on each data point. The estimated importance weights $\hat{\alpha}_{m1}, \ldots, \hat{\alpha}_{mK}$ are sent to the server alongside the local model for global updates. The cluster models' updates are performed similarly as in {\ttfamily CDFL} using the received importance weights. As there are no existing works addressing both asynchrony and soft-clustering concurrently in FL, {\ttfamily FedSoft-Async} serves as the most suitable baseline method; and 2) {\ttfamily Local}. The clients only perform local optimizations and never interact with the server.

In the initialization phase, clients perform computations using the averaged cluster model. Each client possesses a dataset ranging from 500 to 2000 data points, with 40\% to 90\% originating from a primary distribution and the remainder from other clusters. After initialization, clients autonomously and randomly decide when to update their models. 
Each upload-download cycle prompts a client to receive new data, necessitating further model updates. In the experiments presented in Table \ref{table: accuracy-all}, the number of clients is 20 times the number of clusters. The value of staleness threshold $\tau_0$ is set to the number of clients. In the FashionMNIST experiments, on average, each client undergoes 25 model updates; and 20 updates in the CIFAR-100 and MiniImagenet-100 experiments. 
In Digit Recognition experiment, we set the client number as 100 and on average, each client undergoes 40 model updates. For each cluster in above experiments, we randomly choose 3000 samples from the training set to pre-train cluster models and 1000 samples from the test dataset as the proxy dataset on server.
We set parameters in Algorithm \ref{alg:DistributionEstimation & UpdateRaTioCompute} as $\beta_0=0.025$, $ a=10$, and $b=5$, and the regularization parameter in \eqref{client objective} as $\rho=0.1$. Values for other hyper-parameters are given in Appendix \ref{Model Structures.}.

All experiments are conducted on a single machine with two Intel Xeon 6226R CPUs, 384GB of memory, and four NVIDIA 3090 GPUs. Each experiment is repeated 5 times, and the average values and the standard deviations are reported.

\subsection{Results}

\textbf{Accuracy.} Average accuracy results across clients and clusters are shown in Table \ref{table: accuracy-all}. Notably, both {\ttfamily CDFL} and {\ttfamily FedSoft-Async} exhibit significant enhancements in client performance compared to only local training, underscoring the importance of client collaboration. Across most experiments, {\ttfamily CDFL} outperforms {\ttfamily FedSoft-Async} for both clients and clusters, particularly for larger $K$. 

\begin{figure*}[t!]
\begin{center}
\includegraphics[width=\linewidth]{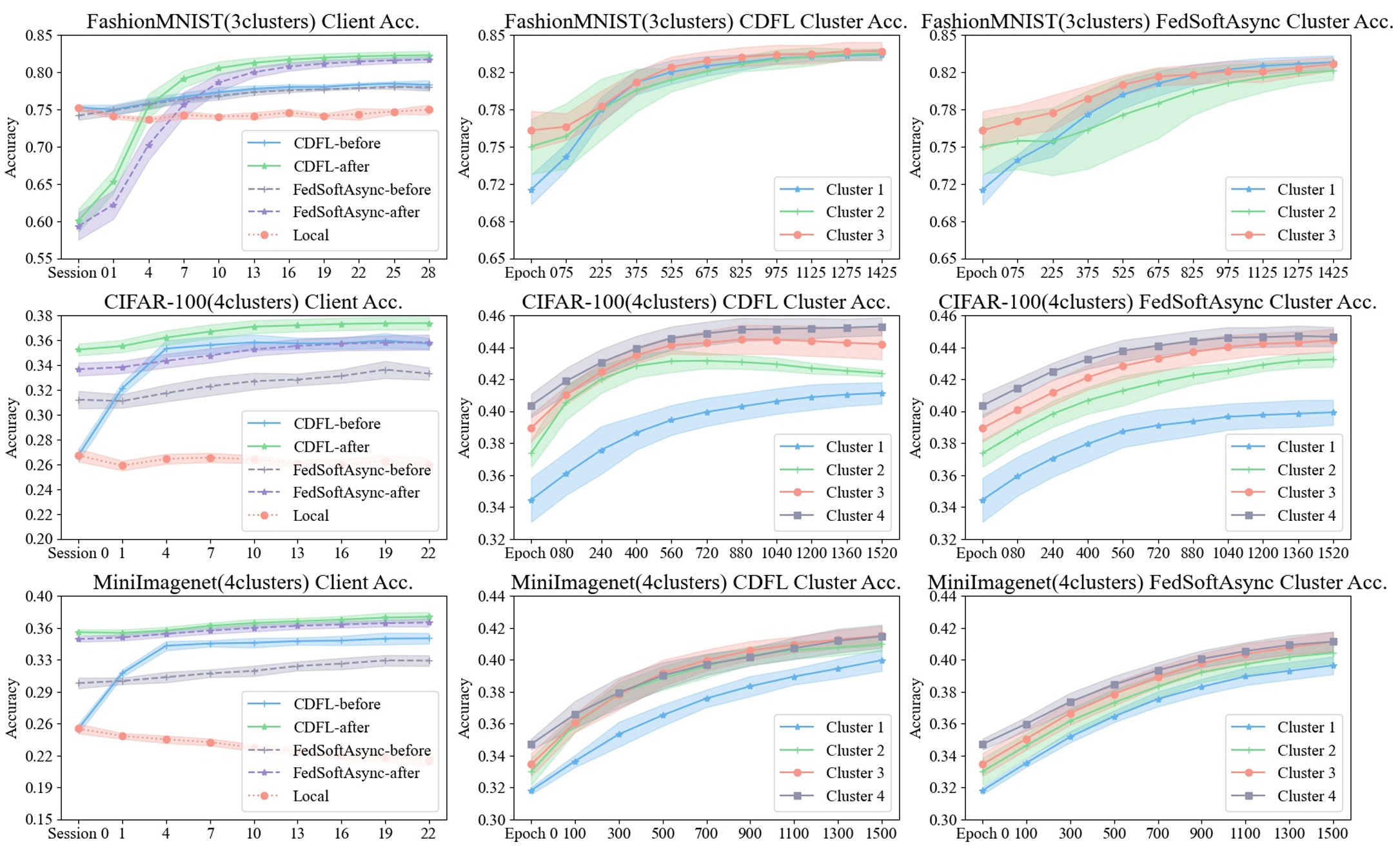}
\vspace{-5mm}
\caption{Average accuracy of clients and clusters over time. 
For client accuracy, in each session all the clients have updated their models for at least once; for cluster accuracy, exact one client updates its model with the sever in each epoch.}
\label{pic:FashionMNIST&MI}
\end{center}
\vspace{-5mm}
\end{figure*}

\begin{figure}[ht!]
\begin{center}
\includegraphics[width=\linewidth]{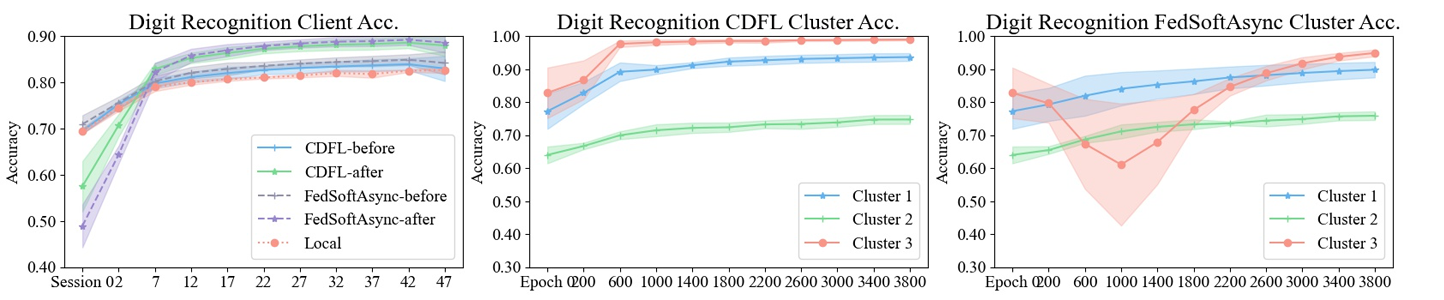}
\vspace{-5mm}
\caption{Average accuracy of clients and clusters on Digit Recognition task over time.}
\label{pic:digit-recognition}
\end{center}
\vspace{-6mm}
\end{figure}

We also plot the client and cluster accuracies over time in~Figure \ref{pic:FashionMNIST&MI}. In FashionMNIST experiments, both {\ttfamily CDFL} and {\ttfamily FedSoft-Async} require a few sessions (each client has updated its model at least once in each session) for the downloaded model to surpass the performance of their locally trained counterparts. This may attribute to the relatively poor performance of the cluster models in the initial epochs. As the accuracies of cluster models improve, the benefit of model updating starts to prevail. Nevertheless, this phenomenon is not observed in CIFAR-100 and MiniImagenet experiments, and model updating with the server helps to boost clients' performance from very early epochs. Additional experiment results on different numbers of clusters can be found in Appendix \ref{more experiments}.
The results for Digit Recognition Task are shown in Table \ref{table: accuracy-all} and in Figure \ref{pic:digit-recognition}. We observe that for both {\ttfamily CDFL} and {\ttfamily FedSoft-Async}, client accuracy before and after model updates have increased. This improvement suggests that clients are receiving better models from the server, enhancing the performance of their local models. While the difference on client accuracy between the proposed algorithm and the baseline is small, {\ttfamily CDFL} has much higher cluster accuracy than the baseline method. 

\begin{figure}[ht!]
\begin{center}
\includegraphics[width=\linewidth]{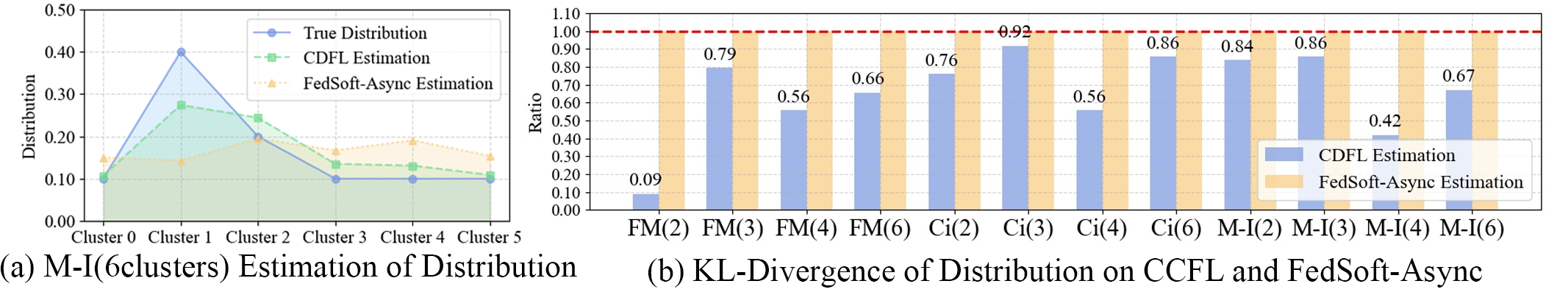}
\vspace{-4mm}
\caption{(a) Estimated distributions in the MiniImagenet (6 clusters) experiment; (b) KL-divergence between the true distribution and the distribution estimated by {\ttfamily CDFL} and {\ttfamily FedSoft-Async}. FM($k$) denotes FashionMNIST ($k$ clusters), Ci as CIFAR-100, M-I as MiniImagenet-100.}
\label{pic:distribution analysis.}
\end{center}
\vspace{-4mm}
\end{figure}

\textbf{Distribution Estimation.}
To assess the proposed distribution estimation method in Algorithm \ref{alg:DistributionEstimation & UpdateRaTioCompute}, we empirically compare the estimation outcomes of {\ttfamily CDFL} and those of {\ttfamily FedSoft-Async}. To quantify this assessment, we employ the KL-divergence metric $KL(P || Q)$, which measures the information loss when approximating the true distribution $P$ with the estimated distribution $Q$. Lower KL divergence value implies more accurate estimation. As shown in Figure~\ref{pic:distribution analysis.}(b), over all datasets and cluster numbers, {\ttfamily CDFL} constantly outperforms {\ttfamily FedSoft-Async}. 

\begin{figure}[ht!]
\begin{center}
\includegraphics[width=\linewidth]{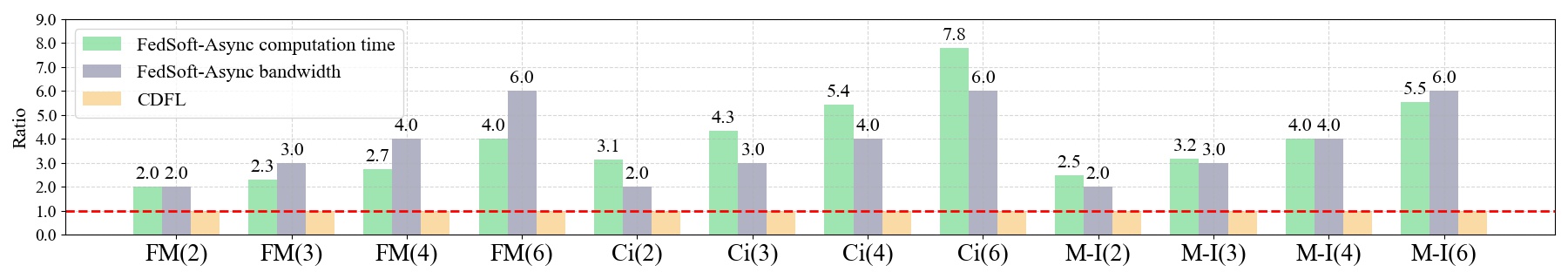}
\caption{Relative communication and computation overheads of {\ttfamily FedSoft-Async} with {\ttfamily CDFL}. 
}
\label{pic:c&c.}
\end{center}
\vspace{-5mm}
\end{figure}

\textbf{Communication and Computation Overhead.} We compare the communication and computation overheads between {\ttfamily FedSoft-Async} and {\ttfamily CDFL} in Figure \ref{pic:c&c.}. Specifically, we focus on download communication overhead, as both methods upload one local model to the server. We normalize both the communication and computation overhead of {\ttfamily CDFL} to 1, and measure the relative values of {\ttfamily FedSoft-Async}. Due to the fact that clients in {\ttfamily CDFL} solely download a single aggregated model, and do not need additional computations beyond local model training, the communication and computation overhead is significantly reduced compared to {\ttfamily FedSoft-Async}. 

\begin{table*}[ht!]
\begin{center}
\caption{Average client and cluster accuracy of FashionMNIST, CIFAR100, and MiniImagenet-100 after final epoch. ``Cli Brf'' and ``Cli Aft'' denote the accuracy of client's model before and after uploading. Average accuracy and standard deviation over 5 trials are reported.}
\label{table: accuracy-all}
\centering
\renewcommand{\arraystretch}{1.2} 
{\fontsize{8}{8.5}\selectfont
\begin{tabular}{p{2.5cm}|c@{\hspace{2.3pt}}c@{\hspace{2.3pt}}c|c@{\hspace{2.3pt}}c@{\hspace{2.3pt}}c|c}
\hline
\multirow{2}{*}{\centering Dataset (Cluster No.)} & \multicolumn{3}{c|}{{\ttfamily CDFL} ACC.} & \multicolumn{3}{c|}{{\ttfamily FedSoft-Async} ACC.} & \multirow{2}{*}{\centering \parbox{2cm}{\centering{\ttfamily Local}\\ \centering Client ACC.}}
 \\
\cline{2-7}
& Cli Bfr & Cli Aft & Cluster & Cli Bfr & Cli Aft & Cluster & \\
\hline
FashionMNIST (2) & .799$\pm$\scriptsize{.011} & .834$\pm$\scriptsize{.003} & .838$\pm$\scriptsize{.007}& .798$\pm$\scriptsize{.012} & .836$\pm$\scriptsize{.003}& .833$\pm$\scriptsize{.008}& .781$\pm$\scriptsize{.019}\\
FashionMNIST (3) & .783$\pm$\scriptsize{.015}& .823$\pm$\scriptsize{.003}& .835$\pm$\scriptsize{.003}&
.780$\pm$\scriptsize{.014}& .818$\pm$\scriptsize{.003}& .823$\pm$\scriptsize{.006}& .746$\pm$\scriptsize{.053}\\
FashionMNIST (4) & .765$\pm$\scriptsize{.020}& .800$\pm$\scriptsize{.006}& .829$\pm$\scriptsize{.005}& 
.759$\pm$\scriptsize{.021}& .786$\pm$\scriptsize{.007}& .800$\pm$\scriptsize{.017}& .694$\pm$\scriptsize{.072}\\
FashionMNIST (6) &.768$\pm$\scriptsize{.020} & .788$\pm$\scriptsize{.018}& .818$\pm$\scriptsize{.010}& .760$\pm$\scriptsize{.024}& .761$\pm$\scriptsize{.023}& .769$\pm$\scriptsize{.051}& .697$\pm$\scriptsize{.074}\\
CIFAR-100 (2) &.370$\pm$\scriptsize{.022} & .392$\pm$\scriptsize{.007}& .417$\pm$\scriptsize{.015}& .369$\pm$\scriptsize{.025}& .397$\pm$\scriptsize{.004}& .416$\pm$\scriptsize{.011}&
.280$\pm$\scriptsize{.029} \\
CIFAR-100 (3) &.284$\pm$\scriptsize{.033} & .303$\pm$\scriptsize{.029}& .362$\pm$\scriptsize{.032}& .274$\pm$\scriptsize{.031}& .295$\pm$\scriptsize{.008}& .348$\pm$\scriptsize{.024}& .208$\pm$\scriptsize{.033}\\
CIFAR-100 (4) & .360$\pm$\scriptsize{.029}& .376$\pm$\scriptsize{.012}&
.432$\pm$\scriptsize{.019}& .337$\pm$\scriptsize{.037}& .361$\pm$\scriptsize{.015}& .430$\pm$\scriptsize{.022}& .261$\pm$\scriptsize{.034}\\
CIFAR-100 (6) &
.293$\pm$\scriptsize{.033}& .304$\pm$\scriptsize{.008}& .376$\pm$\scriptsize{.019}& .267$\pm$\scriptsize{.043}& 
.392$\pm$\scriptsize{.017}& .371$\pm$\scriptsize{.035}&
.214$\pm$\scriptsize{.038} \\
MiniImagenet (2) &
.342$\pm$\scriptsize{.020}& 
.369$\pm$\scriptsize{.003}& .385$\pm$\scriptsize{.006}& 
.340$\pm$\scriptsize{.026}& .374$\pm$\scriptsize{.003}& .388$\pm$\scriptsize{.002}& .225$\pm$\scriptsize{.030}\\
MiniImagenet (3) &
.291$\pm$\scriptsize{.029}&
.314$\pm$\scriptsize{.018}&
.359$\pm$\scriptsize{.008}&
.276$\pm$\scriptsize{.033}&
.308$\pm$\scriptsize{.011}& .353$\pm$\scriptsize{.004}&
.180$\pm$\scriptsize{.028} \\
MiniImagenet (4) &
.351$\pm$\scriptsize{.026}& .376$\pm$\scriptsize{.014}& .411$\pm$\scriptsize{.007}& .328$\pm$\scriptsize{.035}& .371$\pm$\scriptsize{.009}& .407$\pm$\scriptsize{.008}& .219$\pm$\scriptsize{.029} \\
MiniImagenet (6) &
.309$\pm$\scriptsize{.029}&
.334$\pm$\scriptsize{.007}&
.386$\pm$\scriptsize{.011}&
.280$\pm$\scriptsize{.038}&
.323$\pm$\scriptsize{.011}&
.380$\pm$\scriptsize{.013}& 
.192$\pm$\scriptsize{.029}\\
DigitRecognition &
.835$\pm$\scriptsize{.077}&
.883$\pm$\scriptsize{.056}&
.890$\pm$\scriptsize{.130}&
.846$\pm$\scriptsize{.070}&
.890$\pm$\scriptsize{.052}&
.870$\pm$\scriptsize{.102}& 
.820$\pm$\scriptsize{.083}\\

\hline
\end{tabular}
}
\end{center}
\vspace{-3mm}
\end{table*}



\subsection{Ablation Study}
\label{Ablation Study}
\begin{figure}[h!]
\begin{center}
\includegraphics[width=\textwidth]{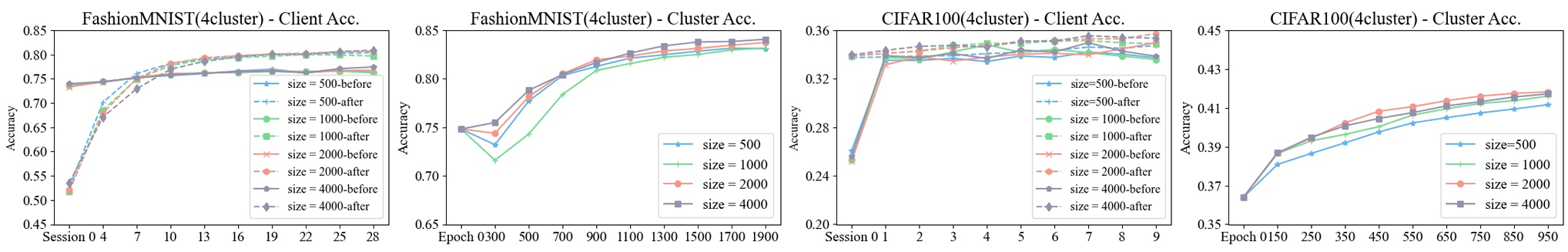}
\end{center}
\caption{Average client and cluster accuracy for different sizes of proxy datasets at the server. 
}
\label{pic:ab-different size of public dataset.}
\vspace{-6mm}
\end{figure}

\textbf{Size of Proxy Dataset.} 
We evaluate the performance of {\ttfamily CDFL} for various sizes of public datasets on FashionMNIST(4 clusters) and CIFAR100(4clusters) in Figure \ref{pic:ab-different size of public dataset.}, ranging from 500 to 4000 samples. In those experiments, although the performance difference for using smaller sizes of proxy datasets has relatively lower performances, but the difference is negligible. We demonstrate that even a proxy dataset size of 500 samples (\textbf{approximately 0.8\% of the training data}) can yield relatively favorable performance outcomes. Therefore, we consider the inclusion of a small but available proxy dataset to be a reasonable and beneficial practice in machine learning research. Other ablation studies on different values of $\rho$, $a$, $b$, and different number of clients can be found in Appendix \ref{more ablation study}.

\section{Conclusion}
\label{conclusion}
\vspace{-3mm}
We introduce Client-Driven Federated Learning ({\ttfamily CDFL}), a novel FL paradigm that emphasizes on clients' autonomy, performance, and efficiency. In {\ttfamily CDFL}, a client autonomously decides when to update its model with the server, to accommodate potential distribution shifts. {\ttfamily CDFL} proposes novel distribution estimation strategies, for the server who hosts multiple cluster models to accurately estimate the distribution of the client, facilitating a rapid adaptation of the client's model to the new tasks. We theoretically and experimentally demonstrate the superiority of {\ttfamily CDFL} 
over baselines.



\bibliography{ref}


\clearpage
\appendix


\section{Data Structures, Model Structures and Hyper-parameters}
\label{Model Structures.}

\begin{wrapfigure}{r}{0.5\textwidth}
    \centering
    \includegraphics[width=0.5\textwidth]{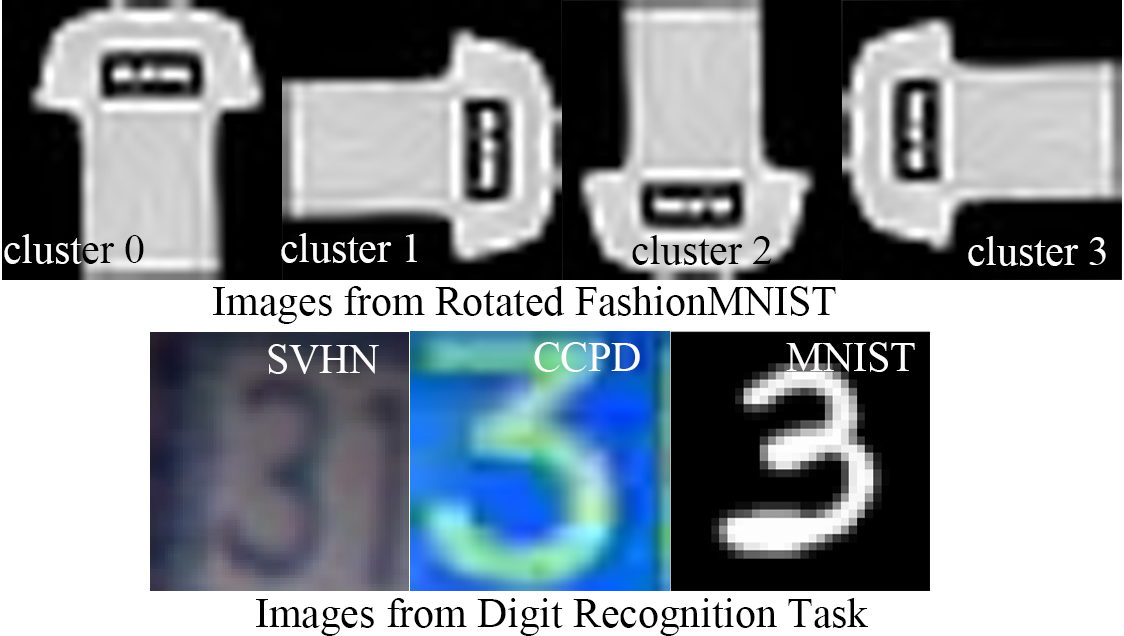}
    \caption{Images from conducted experiments.}
    \label{pic: visualized images}
\end{wrapfigure} 

For the Digit Recognition experiment, the details are as below: \textbf{MNIST\&USPS}: contains 76,000 training images and 14,000 testing images. \textbf{SVHN}: contains 73,257 training images and 26,032 testing images. Numbers from \textbf{CCPD}: contains 80,000 training images and 20,000 testing images. We normalize the all the images to the size of 32*32*3 and employ the CNN model to do the training. All the other settings are similar to those in FashionMNIST experiment. Examples from training data are shown in Figure \ref{pic: visualized images}.

We use CNN models to train FashionMNIST and Digit Recognition task and ResNet18 to train Cifar-100 and MiniImagenet-100. The model structures are as follows.
\begin{itemize}
    \item This CNN model consists of two convolutional layers with 5x5 kernels and ReLU activation, each followed by max-pooling layers with 2x2 kernels. This is followed by two fully-connected layers with 512 and 10 neurons, respectively. For the FashionMNIST task, the model takes 28x28 grayscale images as input; for the digit recognition task, the model takes 32*32*3 umages as input. The model produces class probabilities for 10 classes. It employs ReLU activation throughout.
    \item ResNet18: ResNet-18 is characterized by its residual blocks, and it's a smaller variant of the original ResNet architecture. The model consists of two types of residual blocks: BasicBlock, containing two convolutional layers with 3x3 kernels and ReLU activations, along with batch normalization. It also includes shortcut connections to handle different input and output dimensions when the stride is not equal to 1; BottleNeck, including three convolutional layers with 1x1, 3x3, and 1x1 kernels, along with ReLU activations and batch normalization and uses shortcut connections for dimension matching. The ResNet-18 architecture is structured around an initial convolutional layer with 64 output channels and a 3x3 kernel with padding set to 1, followed by four stages of residual blocks. Each stage contains a specific number of residual blocks, with the configuration typically set as [2, 2, 2, 2].Within each residual block, the block type can be either BasicBlock or BottleNeck. The number of output channels in the convolutional layers is typically set according to the block type: 64 for BasicBlock and 256 for BottleNeck. Stride values are often set to 1 for BasicBlock and 2 for BottleNeck to achieve downsampling. The expansion factor is 1 for BasicBlock and 4 for BottleNeck. The model uses adaptive average pooling to produce a fixed-sized output, typically (1, 1) spatial dimensions, and a fully-connected layer for classification, with the number of output classes of 100 for CIFAR-100 and activation functions serving as essential hyperparameters.
\end{itemize}

For all experiments, batch size is chosen as 128. For CNN model, Adam optimizer is chosen with weight decay of 0.005 and learning rate of 0.01. For ResNet models, SGD opitimizer is chosen with weight decay of 5e-4, momentum of 0.9 and learning rate as 0.1. Cross entropy loss is computed for optimization through the experiments. Distribution estimation and server updating parameters are listed as follows in Table \ref{table: distribution estimation}.

In the experiment, each client is initially assigned a main cluster index before training. During the training's outset, every client is randomly provided with 500-2000 data samples, with 40\%-90\% sourced from the designated main cluster and the remainder from other cluster distributions. Clients conduct local training and autonomously decide when to upload their models. If a client initiates an upload session, the accuracy of the uploaded model is first evaluated on the test sets (consisting of the same distribution as their training sets), termed as "Client Before Accuracy." Subsequently, upon downloading the new personalized model, the accuracy of the received model is assessed as "Client After Accuracy." After each update of every single client, he would receive new data sampled as the same way mentioned above. Since the distribution has changed, the client needs to do training on new data again and decides the next upload.

\begin{table}[h!]
\caption{Distribution estimation and server updating parameters. \textit{min} means to use the minimum of the computed $l_k$ (or $d_{1k}, d_{2k}$) as bar. \textit{ave} means to use the average of the computed $\beta_{1k}$ as bar. If multiple values are resented in $A$, it means to do \textit{softmax} multiple times with respective given value.}
\begin{center}
\label{table: distribution estimation}
\begin{small}
\centering  
\renewcommand{\arraystretch}{1.2} 
\begin{tabular}{c|c c c c c c|c}
\hline
\multirow{2}{*}{\centering Dataset (cluster No.)} & \multicolumn{6}{c}{Distribution Estimation} & \multicolumn{1}{|c}{Server Updating}\\
\cline{2-8}
& \small $c_1$ & \small $c_2$ & \small $A$ & \small $l_{bar}$ & \small $d_{1bar}$ & \small $d_{2bar}$ & \small $\beta_{1bar}$  \\
\hline
FashionMNIST (2) & 0.5 & 0.4 & 3 & 8 & 0& 0& ave \\
FashionMNIST (3) & 0.5 & 0.25 & 3 & 8 & 0& 0& ave \\
FashionMNIST (4) & 0.5 & 0.25 & 7& 8 & 0& 0& ave\\
FashionMNIST (6) & 0.7 & 0.2 & 15& 8 & 0& 0& ave\\
CIFAR-100 (2) & 0.5 & 0.2 & 1 & 40 & 0 & 0 & ave\\
CIFAR-100 (3) & 0.5 & 0.2 & 10 & min & min & min & ave\\
CIFAR-100 (4) & 0.5 & 0.2 & 10 & min & min & min & ave\\
CIFAR-100 (6) & 0.5 & 0.2 & 10 & min & min & min & ave\\
MiniImagenet (2) & 0.5 & 0.2 & 1 & 70 & 0 & 0 & ave\\
MiniImagenet (3) & 0.5 & 0.4 & 7 & min & min & min & ave\\
MiniImagenet (4) & 0.6 & 0.3 & 15 & min & min & min & ave\\
MiniImagenet (6) & 0.6 & 0.3 & 10, 10 & min & min & min & ave\\
DigitRecognition & 0.9 & 0 & 5 & min & min & min & ave\\
\hline
\end{tabular}
\end{small}
\end{center}
\end{table}

\clearpage

\section{Proof of Theorem \ref{theorem 1}}
\label{proof1}
Without the loss of generality, we assume client $m$ uploads $(v_m^t, \tau)$ to the server at epoch $t$. We assume the client is not stale ($t-\tau<\tau_0$), and cluster $k$'s model $w_k^{t-1}$ would be updated with parameter $\beta_{mk}^t > 0$. We assume $v_m^t$ is the result of applying $H_{min} \leq H \leq H_{max}$ local updates to $u_m^\tau$. We define $\mathbb{J}_k(v; u) = F_k(v)+\frac{\rho}{2}\left\|v-u\right\|^2$. For convenience we denote $v_{m,h}^t$ as $v_h$ ($h\in [H]$), $u_m^\tau$ as $u_\tau$, $w_k$ as $w$, $w_k^*$ as $w^*$.

Conditioning on $v_{h-1}$, for $\forall h \in [H]$ we have

\begin{equation}
    \begin{aligned}
        \mathbb{E}[F_k(v_h)-F_k(w^*)] \leq \mathbb{E}[\mathbb{J}_k(v_h; u_\tau)-F_k(w^*)] \leq \mathbb{E}[\mathbb{J}_k(v_h; u_\tau)]-F_k(w^*)
    \end{aligned}
\end{equation}

With the updating function
\begin{equation}
    v_{h} = v_{h-1} -\gamma\nabla J_m(v_{h-1}; u)
\end{equation}

and Taloy's Expansion $f(w-y) = f(x)-\langle\nabla f(x), y \rangle+\frac{1}{2}\nabla^2f(x)y^2$ and using $L_k$-smoothness,

\begin{equation}
\label{eq-proof: taylor expansion and L-smooth}
\begin{aligned}
     & \mathbb{J}_k(v_h; u_\tau) \\
     & = \mathbb{J}_k\left(v_{h-1}-\gamma\nabla J_m(v_{h-1}; u_\tau); u_\tau\right)\\
     & = \mathbb{J}_k(v_{h-1}; u_\tau)-\langle\nabla \mathbb{J}_k(v_{h-1}; u_\tau), \gamma \nabla J_m(v_{h-1}; u_\tau)\rangle \\
     &\qquad\qquad+\frac{1}{2}\nabla^2\mathbb{J}_k(v_{h-1}; u_\tau)\gamma^2\left\|\nabla J_m(v_{h-1}; u_\tau)\right\|^2\\
     & \leq \mathbb{J}_k(v_{h-1}; u_\tau)-\langle\nabla \mathbb{J}_k(v_{h-1}; u_\tau), \gamma \nabla J_m(v_{h-1}; u_\tau)\rangle +\frac{1}{2}L_k\gamma^2\left\|\nabla J_m(v_{h-1}; u_\tau)\right\|^2
\end{aligned}
\end{equation}

Since $\left\|v_{h-1}-u_\tau\right\|^2 \leq H_{max}^2\gamma^2V_2$ and $\left\|J_m(v_{h-1}; u_\tau)\right\|^2 \leq V_2$,  with \eqref{eq-proof: taylor expansion and L-smooth},

\begin{equation}
\label{eq-proof: first leq}
    \begin{aligned}
        & \mathbb{E}[F_k(v_h)-F_k(w^*)] \\
        & \leq \mathbb{J}_k(v_{h-1}; u_\tau)-F_k(w^*)-\gamma\mathbb{E}[\langle\nabla \mathbb{J}_k(v_{h-1}; u_\tau), \nabla J_m(v_{h-1}; u_\tau)\rangle]\\
        & \qquad\qquad+ \frac{1}{2}L_k\gamma^2\mathbb{E}[\left\|\nabla J_m(v_{h-1}; u_\tau)\right\|^2]\\
        & = F_k(v_{h-1})-F_k(w^*)+\frac{\rho}{2}\left\|v_{h-1}-u_\tau\right\|^2-\gamma\mathbb{E}[\langle\nabla \mathbb{J}_k(v_{h-1}; u_\tau), \nabla J_m(v_{h-1}; u_\tau)\rangle]\\
        & \qquad\qquad+ \frac{1}{2}L_k\gamma^2\mathbb{E}[\left\|\nabla J_m(v_{h-1}; u_\tau)\right\|^2]\\
        & \leq F_k(v_{h-1})-F_k(w^*)-\gamma\mathbb{E}[\langle\nabla \mathbb{J}_k(v_{h-1}; u_\tau), \nabla J_m(v_{h-1}; u_\tau)\rangle]+\frac{L_k\gamma^2}{2}V_2+\frac{\rho H_{max}^2\gamma^2}{2}V_2
    \end{aligned}
\end{equation}

Take a small constant $\epsilon > 0$, and with inequality of arithmetic and geometric mean, if we choose some $\rho \geq  \frac{2V_1+\frac{1}{2}\left\|v_{h-1}-u_\tau\right\|^2+\sqrt{4\left\|v_{h-1}-u_\tau\right\|^2(1+V_1)\epsilon}}{2\left\|v_{h-1}-u_\tau\right\|^2}$ for all possible $v_{h-1}, u_\tau$, we have

\begin{equation}
    \begin{aligned}
        &\langle\nabla \mathbb{J}_k(v_{h-1}; u_\tau), \nabla J_m(v_{h-1}; u_\tau)\rangle-\epsilon\left\|\nabla F_k(v_{h-1})\right\|^2\\
        & = \langle \nabla F_k(v_{h-1})+\rho(v_{h-1}-u_\tau), \nabla f_m(v_{h-1})+\rho(v_{h-1}-u_\tau)\rangle-\epsilon\left\|\nabla F_k(v_{h-1})\right\|^2\\
        & = \langle \nabla F_k(v_{h-1}), \nabla f_m(v_{h-1}) \rangle+\rho\langle \nabla F_k(v_{h-1})+\nabla f_m(v_{h-1}), v_{h-1}-u_\tau \rangle\\
        & \quad\quad +\rho^2\left\|v_{h-1}-u_\tau\right\|^2-\epsilon\left\|\nabla F_k(v_{h-1})\right\|^2\\
        & \geq -\frac{1}{2}\left\|\nabla F_k(v_{h-1})\right\|^2-\frac{1}{2}\left\|\nabla f_m(v_{h-1})\right\|^2-\frac{\rho}{2}\left\|\nabla F_k(v_{h-1})+f_m(v_{h-1})\right\|^2\\
        & \quad\quad -\frac{\rho}{2}\left\|v_{h-1}-u_\tau\right\|^2+\rho^2\left\|v_{h-1}-u_\tau\right\|^2-\epsilon\left\|\nabla F_k(v_{h-1})\right\|^2\\
        & \geq -\frac{1}{2}\left\|\nabla F_k(v_{h-1})\right\|^2-\frac{1}{2}\left\|\nabla f_m(v_{h-1})\right\|^2-\rho\left\|\nabla F_k(v_{h-1})\right\|^2-\rho\left\|\nabla f_m(v_{h-1})\right\|^2\\
        & \quad\quad -\frac{\rho}{2}\left\|v_{h-1}-u_\tau\right\|^2+\rho^2\left\|v_{h-1}-u_\tau\right\|^2-\epsilon\left\|\nabla F_k(v_{h-1})\right\|^2\\
        & = \left\|v_{h-1}-u_\tau\right\|^2\rho^2-\left(\left\|\nabla F_k(v_{h-1})\right\|^2+\left\|\nabla f_m(v_{h-1})\right\|^2+\frac{1}{2}\left\|v_{h-1}-u_\tau\right\|^2\right)\rho\\
        & \quad\quad -\left(\frac{1}{2}\left\|\nabla F_k(v_{h-1})\right\|^2+\frac{1}{2}\left\|\nabla f_m(v_{h-1})\right\|^2+\epsilon\left\|\nabla F_k(v_{h-1})\right\|^2\right)\\
        & \geq \left\|v_{h-1}-u_\tau\right\|^2\rho^2-\left(2V_1+\frac{1}{2}\left\|v_{h-1}-u_\tau\right\|^2\right)\rho-(1+V_1)\epsilon \geq 0
    \end{aligned}
\end{equation}

Thus, we have
\begin{equation}
\label{eq-proof: help-eq1}
    \gamma\langle\nabla \mathbb{J}_k(v_{h-1}; u_\tau), \nabla J_m(v_{h-1}; u_\tau)\rangle\geq\gamma\epsilon\left\|\nabla F_k(v_{h-1})\right\|^2
\end{equation}

Taking \eqref{eq-proof: help-eq1} into \eqref{eq-proof: first leq}, we have

\begin{equation}
\label{eq-proof: iteration1}
    \mathbb{E}[F_k(v_h)-F_k(w^*)] \leq F_k(v_{h-1})-F_k(w^*)-\gamma\epsilon\left\|\nabla F_k(v_{h-1})\right\|^2+\frac{L_k\gamma^2}{2}V_2+\frac{\rho H_{max}^2\gamma^2}{2}V_2
\end{equation}

By iterating \eqref{eq-proof: iteration1} for $h = 0, ..., H-1$, we have

\begin{equation}
    \mathbb{E}[F_k(v_h)-F_k(v_0)] \leq -\gamma\epsilon\sum_{h=0}^{H-1}\left\|\nabla F_k(v_h)\right\|^2+\frac{H_{max}L_k\gamma^2}{2}V_2+\frac{\rho H_{max}^3\gamma^2}{2}V_2
\end{equation}

Since $v_0$ is initiated from $u_\tau$, we can rewrite above equation as 

\begin{equation}
\label{eq-proof: compo1}
    \mathbb{E}[F_k(v_h)-F_k(u_\tau)] \leq -\gamma\epsilon\sum_{h=0}^{H-1}\left\|\nabla F_k(v_h)\right\|^2+\frac{H_{max}L\gamma^2}{2}V_2+\frac{\rho H_{max}^3\gamma^2}{2}V_2
\end{equation}

We know that cluster $k$ is not updated in every iteration $t$. If we relabel the iterations when cluster $k$ is updated as $0, 1, 2, ..., s-1, s=t$, then we have $w_k^s = (1-\beta_{mk}^s)w_k^{s-1}+\beta_{mk}^s v_m^s$. For simplicity, we denote $w_k^s$ as $w_s$, $\beta_{mk}^s$ as $\beta_s$, then

\begin{equation}
\label{eq-proof:mid1}
    \begin{aligned}
        & \mathbb{E}[F_k(w_s)-F_k(w_{s-1})]\\
        & \leq \mathbb{E}[\mathbb{J}_k(w_s; w_{s-1})-F_k(w_{s-1})]\\
        & = \mathbb{E}[\mathbb{J}_k((1-\beta_s)w_{s-1}+\beta_s v_m^s; w_{s-1})-F_k(w_{s-1})]\\
        & = \mathbb{E}[\mathbb{J}_k((1-\beta_s)w_{s-1}+\beta_s v_h; w_{s-1})-F_k(w_{s-1})]\\
        & \leq \mathbb{E}[(1-\beta_s)\mathbb{J}_k(w_{s-1}; w_{s-1})+\beta_s\mathbb{J}_k(v_h; w_{s-1})-F_k(w_{s-1})]\\
        & = \mathbb{E}[\beta_s\left(F_k(v_h)-F(w_{s-1})\right)+\frac{\beta_s\rho}{2}\left\|v_h-w_{s-1}\right\|^2]\\
        & \leq \beta_s \mathbb{E}[F_k(v_h)-F_k(w_{s-1})]+\beta_s\rho\left\|v_h-u_\tau\right\|^2+\beta_s\rho\left\|u_\tau-w_{s-1}\right\|^2
    \end{aligned}
\end{equation}

We have

\begin{equation}
\begin{aligned}
    & \left\|u_\tau-w_{s-1}\right\| \\
    & =  \left\|u_m^\tau-w_k^{s-1}\right\|\\
    &=\left\|\sum_{i=1}^K\hat{\alpha}_{mi}^s w_i^{\tau}-w_k^{s-1}\right\|\\
    &=\left\|\sum_{i=1}^K \hat{\alpha}_{mi}^s w_i^\tau-w_k^{\tau}+w_k^{\tau}-w_k^{s-1}\right\|\\
    &=\left\|\sum_{i=1}^K\hat{\alpha}_{mi}^s\left( w_i^\tau-w_k^{\tau}\right)+\left(w_k^{\tau}-w_k^{s-1}\right)\right\|\\
    &\leq \sum_{i=1}^K \hat{\alpha}_{mi}^s\left\| w_i^\tau-w_k^{\tau}\right\|+\left\|w_k^{\tau}-w_k^{s-1}\right\|\\
    &\leq \sum_{i=1}^K \left\| w_i^\tau-w_k^{\tau}\right\|+\left\|w_k^{\tau}-w_k^{s-1}\right\|\\
\end{aligned}
\end{equation}

We can notice that

\begin{equation}
    \begin{aligned}
        &\left\| w_i^\tau-w_k^\tau\right\|\\
        & = \left\| w_i^\tau-w_i^*+w_i^*-w_k^*+w_k^*-w_k^\tau\right\|\\
        & \leq \left\| w_i^\tau-w_i^*\right\|+\left\|w_i^*-w_k^*\right\|+\left\|w_k^*-w_k^\tau\right\|
    \end{aligned}
\end{equation}

Since $\left\|w_k\right\| \leq a_k \Delta$, then $\left\|w_k-w_k^*\right\| \leq 2a_k \Delta$, thus
\begin{equation}
    \left\|w_i^\tau-w_k^\tau\right\|\leq 2a_i\Delta+\Delta+2a_k\Delta = (2a_i+2a_k+1)\Delta
\end{equation}

And,
\begin{equation}
\begin{aligned}
    & \left\|w_k^\tau-w_k^{s-1}\right\| = \left\|w_k^\tau-w_k^*+w_k^*-w_k^{s-1}\right\| \\
    & \leq \left\|w_k^\tau-w_k^*\right\|+\left\|w^*-w_k^{s-1}\right\| \leq 2a_k\Delta+2a_k\Delta=4a_k\Delta
\end{aligned}
\end{equation}

Thus,
\begin{equation}
    \left\|u_\tau-w_{s-1}\right\| \leq \left(2\sum_{i=1}^Ka_i+(2K+1)a_k+K\right)\Delta
\end{equation}

And with $\left\|v_h-u_\tau\right\|^2 \leq H^2_{max}\gamma^2V_2$, from \eqref{eq-proof:mid1} we have

\begin{equation}
\label{eq-proof: mid2}
    \begin{aligned}
        & \mathbb{E}[F_k(w_s)-F_k(w_{s-1})]\\
        & \leq \beta_s \mathbb{E}[F_k(v_h)-F_k(w_{s-1})]+\beta_s\rho H^2_{max}\gamma^2V_2+\beta_s\rho\left(2\sum_{i=1}^Ka_i+(2K+1)a_k+K\right)^2\Delta^2\\
        & \leq \beta_s \mathbb{E}[F_k(v_h)-F_k(u_\tau)+F_k(u_\tau)-F_k(w_{s-1})]+\beta_s\rho H^2_{max}\gamma^2V_2\\
        &\qquad+\beta_s\rho\left(2\sum_{i=1}^Ka_i+(2K+1)a_k+K\right)^2\Delta^2
    \end{aligned}
\end{equation}

Using $L_k$-smoothness, we have

\begin{equation}
\label{eq-prrof: compo2}
\begin{aligned}
    & \mathbb{E}[F_k(u_\tau)-F_k(w_{s-1})]\\
    & \leq \langle \nabla F_k(w_{s-1}), u_\tau-w_{s-1}\rangle+\frac{L_k}{2}\left\|u_\tau-w_{s-1}\right\|^2\\
    & \leq \left\|F_k(w_{s-1})\right\|\left\|u_\tau-w_{s-1}\right\|+\frac{L_k}{2}\left\|u_\tau-w_{s-1}\right\|^2\\
    & \leq \sqrt{V_1} \left(2\sum_{i=1}^Ka_i+(2K+1)a_k+K\right)\Delta+\frac{L_k}{2}\left(2\sum_{i=1}^Ka_i+(2K+1)a_k+K\right)^2\Delta^2
\end{aligned}
\end{equation}

With the inequalities from \eqref{eq-proof: compo1}, \eqref{eq-prrof: compo2}, we can rewrite \eqref{eq-proof: mid2} into

\begin{equation}
\label{eq-proof: mid3}
    \begin{aligned}
        & \mathbb{E}[F_k(w_s)-F_k(w_{s-1})]\\
        & \leq -\beta_s\gamma\epsilon\sum_{h=0}^{H-1}\left\|\nabla F_k(v_h)\right\|^2+\frac{H_{max}L_k\gamma^2}{2}\beta_s V_2+\frac{\rho H_{max}^3\gamma^2}{2}\beta_sV_2\\
        &\qquad+\beta_s\sqrt{V_1} \left(2\sum_{i=1}^Ka_i+(2K+1)a_k+K\right)\Delta+\frac{\beta_sL}{2}\left(2\sum_{i=1}^Ka_i+(2K+1)a_k+K\right)^2\Delta^2\\
        &\qquad+\beta_s\rho H^2_{max}\gamma^2V_2+\beta_s\rho\left(2\sum_{i=1}^Ka_i+(2K+1)a_k+K\right)^2\Delta^2\\
        & = -\beta_s\gamma\epsilon\sum_{h=0}^{H-1}\left\|\nabla F_k(v_h)\right\|^2+\left(\frac{L_k}{2}+\rho H_{max}+\frac{\rho H_{max}^2}{2}\right)\gamma^2H_{max}\beta_sV_2\\
        & \qquad +\beta_s\sqrt{V_1} \left(2\sum_{i=1}^Ka_i+(2K+1)a_k+K\right)\Delta\\
        & \qquad +\beta_s\left(\frac{L}{2}+\rho\right)\left(2\sum_{i=1}^Ka_i+(2K+1)a_k+K\right)^2\Delta^2
    \end{aligned}
\end{equation}

Denoting $H_s$ as the number of the local iterations applied on client $m$ before be uploads at epoch $s$, by rearranging terms in \eqref{eq-proof: mid3}, we have

\begin{equation}
\label{eq-proof: mid4}
    \begin{aligned}
        & \sum_{h=0}^{H_s-1}\left\|\nabla F_k(v_h)\right\|^2\\
        &\leq \frac{\mathbb{E}[F_k(w_{s-1})-F_k(w_{s})]}{\beta_s\gamma\epsilon}+\frac{\left(\frac{L_k}{2}+\rho H_{max}+\frac{\rho H_{max}^2}{2}\right)\gamma H_{max}V_2}{\epsilon}\\
        & \qquad +\frac{\sqrt{V_1} \left(2\sum_{i=1}^Ka_i+(2K+1)a_k+K\right)\Delta}{\gamma\epsilon}\\
        & \qquad+\frac{\left(\frac{L}{2}+\rho\right)\left(2\sum_{i=1}^Ka_i+(2K+1)a_k+K\right)^2\Delta^2}{\gamma\epsilon}
    \end{aligned}
\end{equation}

Denote $\tau_s$ as if client $m$ uploads at epoch $s$, the last time of client's communication (client last uploads at epoch $\tau_s$ before $s$), by taking total expectation, after $S_k$ global epoch on cluster $k$, where $S_k$ is the total number of validate updates on cluster $k$, we have

\begin{equation}
    \begin{aligned}
        &\mathbb{E}[\left\|\nabla F_k(v)\right\|^2] \\
        & = \mathbb{E}_{\tau_s, s \in \{0, ..., T-1\}, h \in \{0, ..., H'_t-1\}} [\left\|\nabla  F(v_{\tau_s, h}) \right\|^2]\\
        & = \frac{1}{\sum_{s=1}^{S_k}H_s}\sum_{s=1}^{S_k}\sum_{h=0}^{H_s-1}\left\|\nabla F_k(v_h)\right\|^2\\
        &\leq \frac{\mathbb{E}[F_k(w^{0}_k)-F_k(w^{S_k}_k)]}{\beta_0\gamma\epsilon S_k H_{min}}+\frac{\left(\frac{L}{2}+\rho H_{max}+\frac{\rho H_{max}^2}{2}\right)\gamma H_{max}V_2}{\epsilon H_{min}}\\
        & \qquad +\frac{\sqrt{V_1} \left(2\sum_{i=1}^Ka_i+(2K+1)a_k+K\right)\Delta}{\gamma\epsilon H_{min}}\\
        & \qquad+\frac{\left(\frac{L_k}{2}+\rho\right)\left(2\sum_{i=1}^Ka_i+(2K+1)a_k+K\right)^2\Delta^2}{\gamma\epsilon H_{min}}
    \end{aligned}
\end{equation}


\clearpage
\section{Additional Experiment Results}
\label{more experiments}

\begin{figure}[ht!]
\begin{center}
\includegraphics[width=\linewidth]{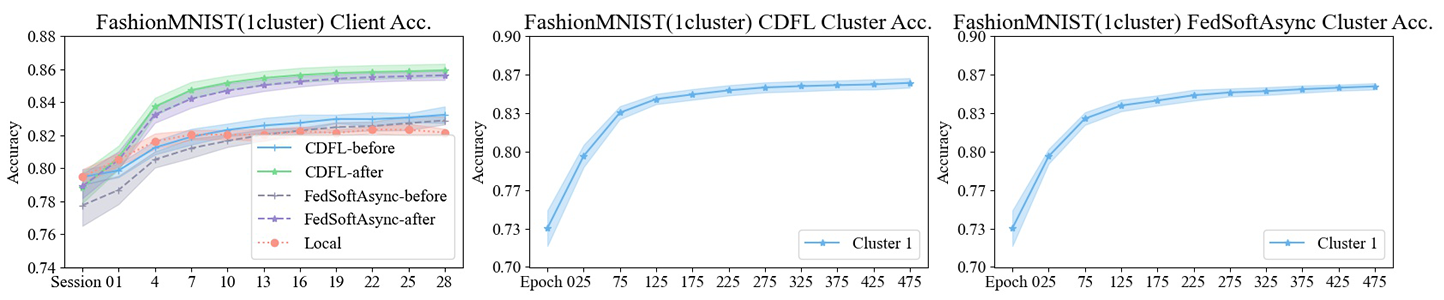}
\caption{Timeflow accuracy of clients and clusters on FashionMNIST (1cluster). Average accuracy of clients is shown for equal times of upload-download cycles.}
\label{pic:FashionMNIST(1cluster)}
\end{center}
\end{figure}

\begin{figure}[ht!]
\begin{center}
\includegraphics[width=\linewidth]{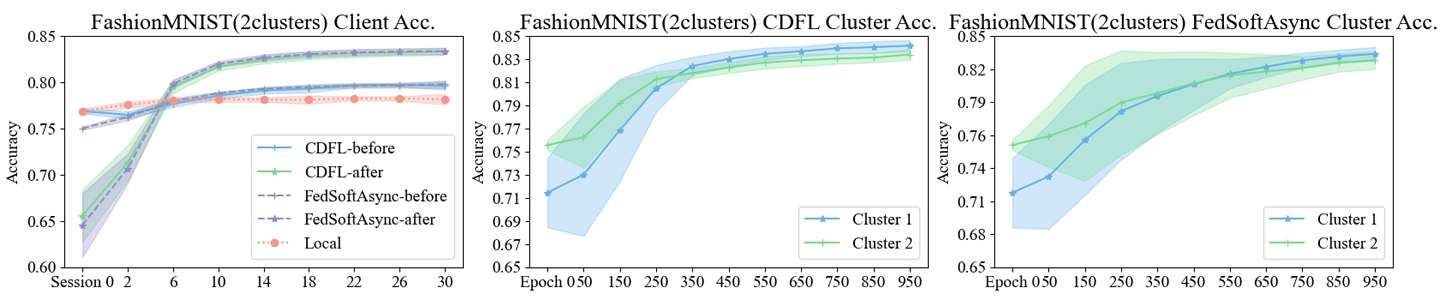}
\caption{Timeflow accuracy of clients and clusters on FashionMNIST (2clusters). Average accuracy of clients is shown for equal times of upload-download cycles.}
\label{pic:FashionMNIST(2clusters)}
\end{center}
\end{figure}

\begin{figure}[ht!]
\begin{center}
\includegraphics[width=\linewidth]{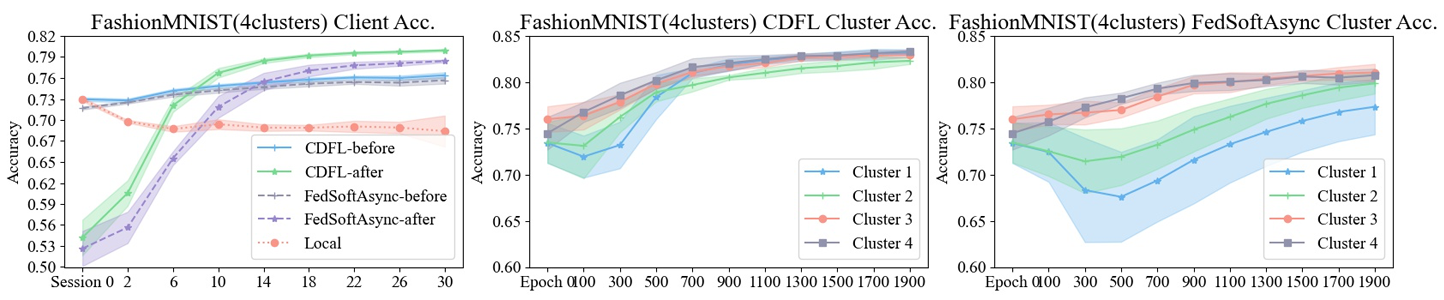}
\caption{Timeflow accuracy of clients and clusters on FashionMNIST (4clusters). Average accuracy of clients is shown for equal times of upload-download cycles.}
\label{pic:FashionMNIST(4clusters)}
\end{center}
\end{figure}

\begin{figure}[ht!]
\begin{center}
\includegraphics[width=\linewidth]{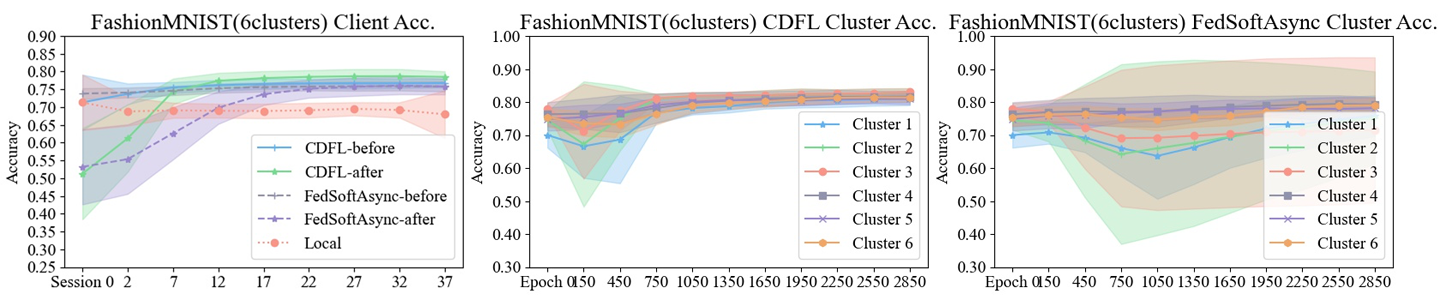}
\caption{Timeflow accuracy of clients and clusters on FashionMNIST (6clusters). Average accuracy of clients is shown for equal times of upload-download cycles.}
\label{pic:FashionMNIST(6clusters)}
\end{center}
\end{figure}

\begin{figure}[ht!]
\begin{center}
\includegraphics[width=\linewidth]{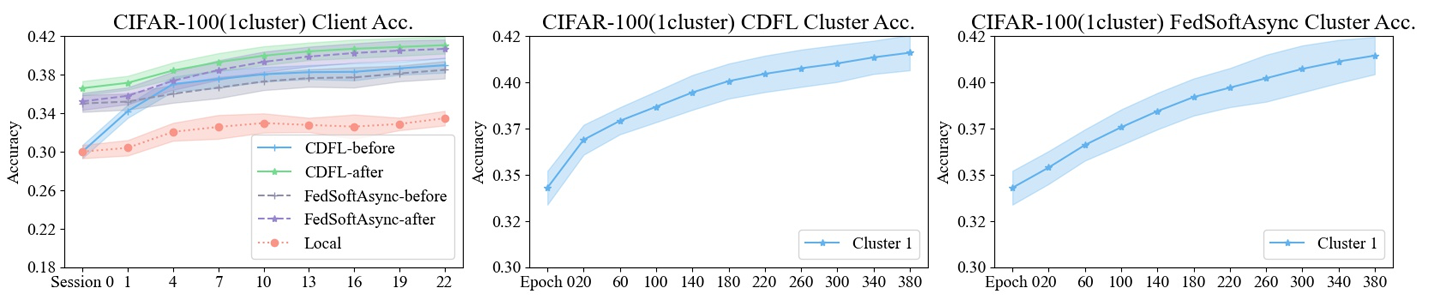}
\caption{Timeflow accuracy of clients and clusters on CIFAR-100 (1cluster). Average accuracy of clients is shown for equal times of upload-download cycles.}
\label{pic:Cifar100(1cluster)}
\end{center}
\end{figure}

\begin{figure}[ht!]
\begin{center}
\includegraphics[width=\linewidth]{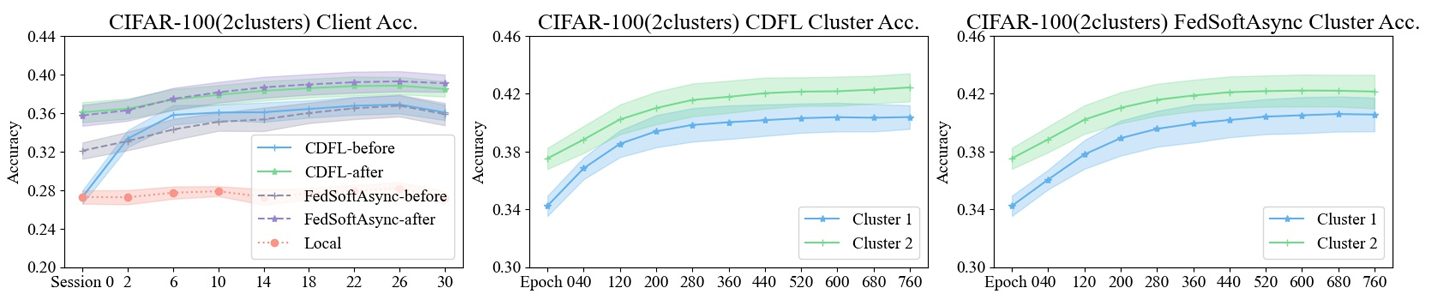}
\caption{Timeflow accuracy of clients and cluseters on CIFAR-100 (2clusters). Average accuracy of clients is shown for equal times of upload-download cycles.}
\label{pic:Cifar100(2clusters)}
\end{center}
\end{figure}

\begin{figure}[ht!]
\begin{center}
\includegraphics[width=\linewidth]{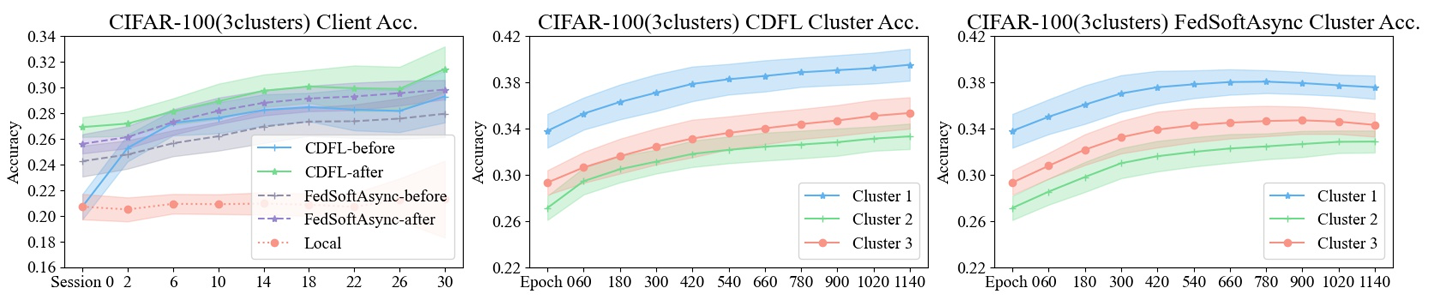}
\caption{Timeflow accuracy of clients and clusters on CIFAR-100 (3clusters). Average accuracy of clients is shown for equal times of upload-download cycles.}
\label{pic:Cifar100(3clusters)}
\end{center}
\end{figure}


\begin{figure}[ht!]
\begin{center}
\includegraphics[width=\linewidth]{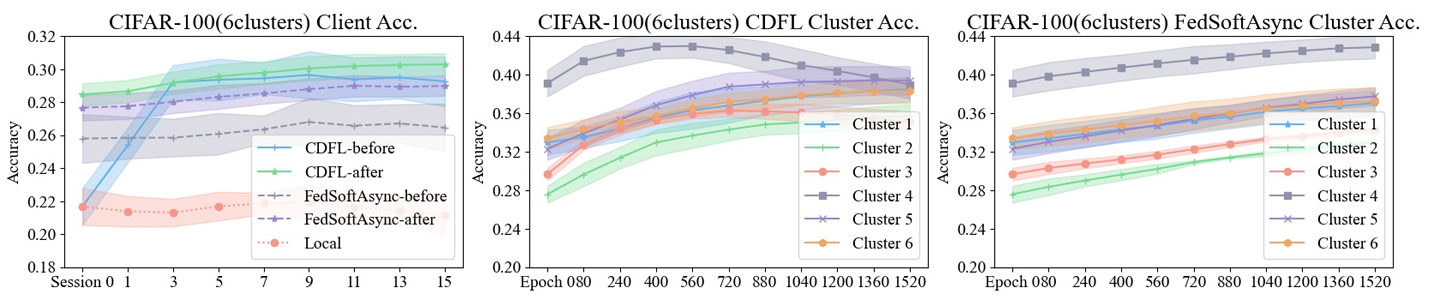}
\caption{Timeflow accuracy of clients and clusters on CIFAR-100 (6clusters). Average accuracy of clients is shown for equal times of upload-download cycles.}
\label{pic:Cifar100(6clusters)}
\end{center}
\end{figure}

\begin{figure}[ht!]
\begin{center}
\includegraphics[width=\linewidth]{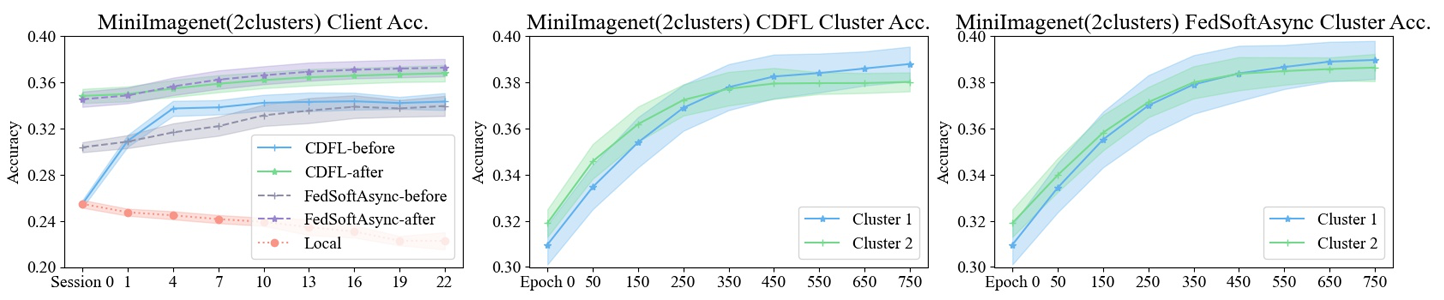}
\caption{Timeflow accuracy of clients and clusters on MiniImagenet-100(2clusters). Average accuracy of clients is shown for equal times of upload-download cycles.}
\label{pic:MiniImagenet(2clusters)}
\end{center}
\end{figure}

\begin{figure}[ht!]
\begin{center}
\includegraphics[width=\linewidth]{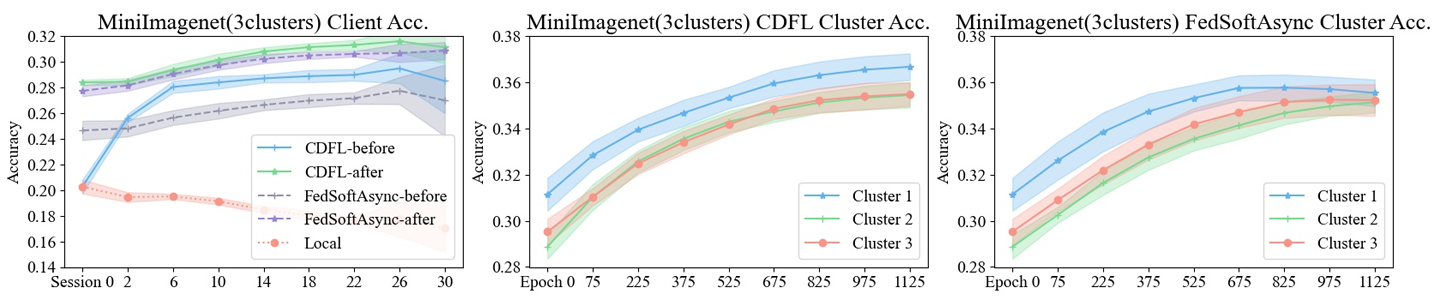}
\caption{Timeflow accuracy of clients and clusters on MiniImagenet-100 (3clusters). Average accuracy of clients is shown for equal times of upload-download cycles.}
\label{pic:MiniImagenet(3clusters)}
\end{center}
\end{figure}

\begin{figure}[ht!]
\begin{center}
\includegraphics[width=\linewidth]{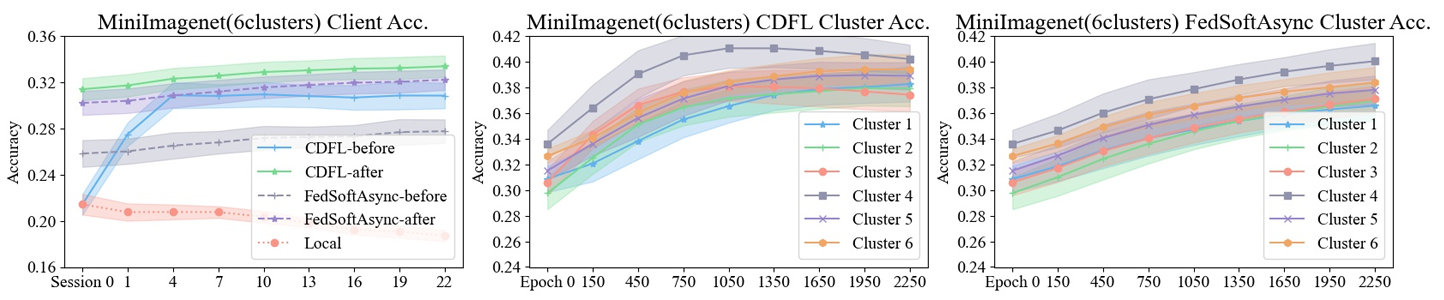}
\caption{Timeflow accuracy of clients and clusters on MiniImagenet-100 (6clusters). Average accuracy of clients is shown for equal times of upload-download cycles.}
\label{pic:MiniImagenet(6clusters)}
\end{center}
\end{figure}

\clearpage
\subsection{More Ablation Studies}
\label{more ablation study}

We further conduct ablation studies on different sets of hyper-parameters in experiments. All the ablation experiments are done within 4 clusters and 100 clients undergoing 2000 global epochs (FashionMNIST) and within 4 clusters and 100 clients undergoing 1000 global epochs (CIFAR-100). 

\begin{figure*}[ht!]
\begin{center}
\includegraphics[width=\linewidth]{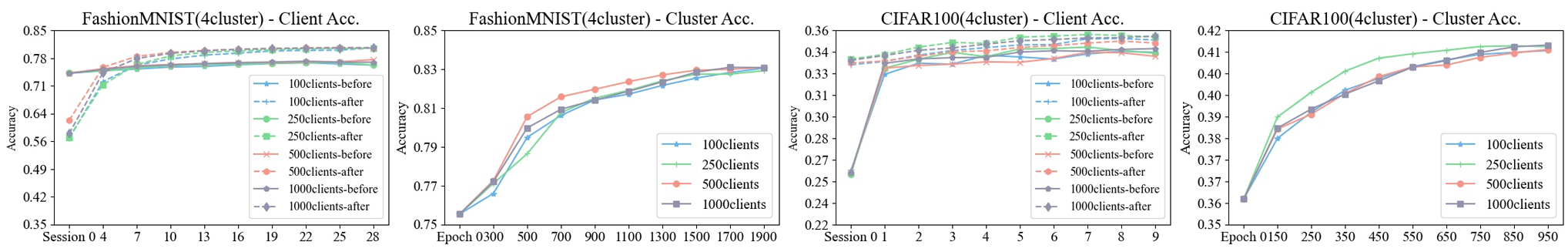}
\caption{Client and Cluster Accuracies of Different Number of Clients on FashionMNIST(4 clusters) and CIFAR100(4 clusters)
}
\label{pic:ablation-study:different-client-number}
\end{center}
\end{figure*}

\textbf{Number of Clients.} We conduct experiments with varying numbers of clients of 100, 250, 500, 1000 on FashionMNIST(4clusters) and CIFAR100(4clusters) and in Figure \ref{pic:ablation-study:different-client-number}. Remarkably, the average accuracy of both clients and clusters exhibited minimal variation across different client counts. This observation underscores the robustness of our system. 

\begin{figure*}[ht!]
\begin{center}
\includegraphics[width=\linewidth]{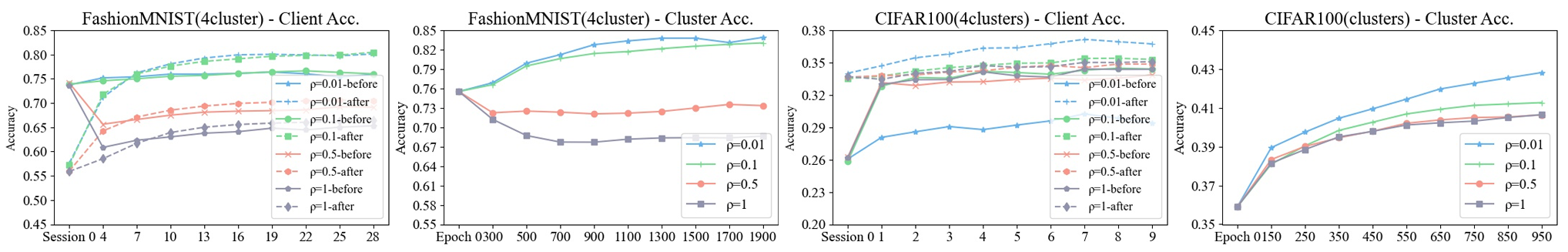}
\caption{Client and Cluster Accuracies of Different $\rho$ on FashionMNIST(4 clusters) and CIFAR100(4 clusters)
}
\label{pic:ablation-study:different-mu}
\end{center}
\end{figure*}

\textbf{Value of $\rho$.} We experiment with the value of the $\rho$ set to 0.01, 0.1, 0.5, and 1 on FashionMNIST(4clusters) CIFAR100(4clusters) and in Figure \ref{pic:ablation-study:different-mu}. The results suggest that on both datasets, smaller $\rho$ leads to better cluster accuracy. However, smaller $\rho$ values, as observed in CIFAR-100 experiments, lead to reduced accuracy on clients' local models before uploading. This may result from the limited amount of local data, and an underutilization of the cluster models with a smaller weight on the regularization term.

\begin{figure*}[ht!]
\begin{center}
\includegraphics[width=\linewidth]{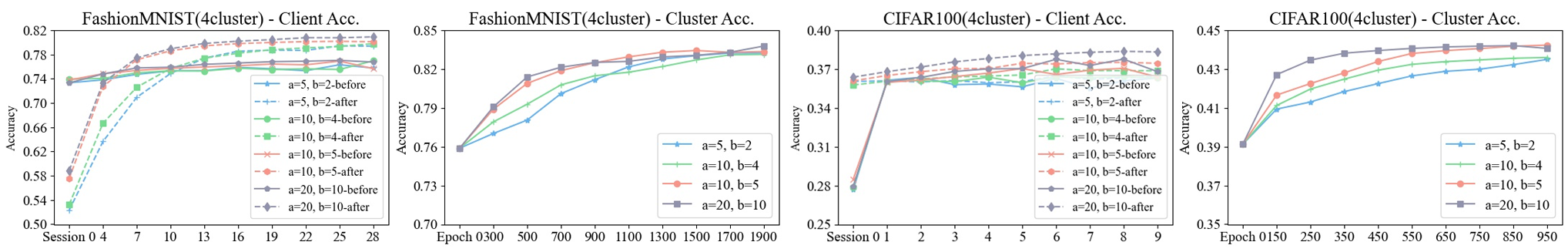}
\caption{Client and Cluster Accuracies of Different $a$ and $b$ on FasionMNIST(4 clusters) and CIFAR100(4 clusters)
}
\label{pic:ablation-study:different-a-and-b}
\end{center}
\end{figure*}

\textbf{Different $a$ and $b$} We have added experiments on different choice of $a$ and $b$ in Algorithm \ref{alg:DistributionEstimation & UpdateRaTioCompute} on FashionMNIST(4clusters) and CIFAR100(4 clusters) and in Figure \ref{pic:ablation-study:different-a-and-b}. From the experiment, we can see the influence of the different combination of $a$ and $b$.

\end{document}